\documentclass[10pt]{article} 
\usepackage[accepted]{tmlr}

\usepackage{amsmath,amsfonts,bm}

\def\eqref#1{equation~\ref{#1}}

\def\1{\bm{1}}

\def\ra{{\textnormal{a}}}

\def\rg{{\textnormal{g}}}

\def\rr{{\textnormal{r}}}

\def\rx{{\textnormal{x}}}

\def\rva{{\mathbf{a}}}

\def\rvu{{\mathbf{i}}}

\def\rvu{{\mathbf{u}}}

\def\rvx{{\mathbf{x}}}

\def\erva{{\textnormal{a}}}

\def\ervx{{\textnormal{x}}}

\def\rmA{{\mathbf{A}}}

\def\vmu{{\bm{\mu}}}
\def\vtheta{{\bm{\theta}}}
\def\va{{\bm{a}}}

\def\vc{{\bm{c}}}

\def\ve{{\bm{e}}}

\def\vn{{\bm{n}}}
\def\vo{{\bm{o}}}

\def\vs{{\bm{s}}}

\def\vu{{\bm{u}}}

\def\vw{{\bm{w}}}
\def\vx{{\bm{x}}}

\def\vz{{\bm{z}}}
\def\vdx{{\bm{dx}}}
\def\vdu{{\bm{du}}}

\def\eva{{a}}

\def\mA{{\bm{A}}}
\def\mB{{\bm{B}}}

\def\mH{{\bm{H}}}
\def\mI{{\bm{I}}}
\def\mJ{{\bm{J}}}

\def\mX{{\bm{X}}}

\def\mSigma{{\bm{\Sigma}}}

\DeclareMathAlphabet{\mathsfit}{\encodingdefault}{\sfdefault}{m}{sl}
\SetMathAlphabet{\mathsfit}{bold}{\encodingdefault}{\sfdefault}{bx}{n}
\newcommand{\tens}[1]{\bm{\mathsfit{#1}}}
\def\tA{{\tens{A}}}

\def\tX{{\tens{X}}}

\def\gG{{\mathcal{G}}}

\def\sA{{\mathbb{A}}}
\def\sB{{\mathbb{B}}}
\def\sC{{\mathbb{C}}}

\def\sP{{\mathbb{P}}}

\def\sR{{\mathbb{R}}}
\def\sS{{\mathbb{S}}}

\def\sU{{\mathbb{U}}}

\def\sW{{\mathbb{W}}}
\def\sX{{\mathbb{X}}}

\def\emA{{A}}

\newcommand{\etens}[1]{\mathsfit{#1}}

\def\etA{{\etens{A}}}

\newcommand{\E}{\mathbb{E}}

\newcommand{\R}{\mathbb{R}}

\newcommand{\KL}{D_{\mathrm{KL}}}
\newcommand{\Var}{\mathrm{Var}}

\newcommand{\Cov}{\mathrm{Cov}}

\newcommand{\normltwo}{L^2}
\newcommand{\normlp}{L^p}

\newcommand{\parents}{Pa} 

\DeclareMathOperator*{\argmax}{arg\,max}
\DeclareMathOperator*{\argmin}{arg\,min}

\usepackage{booktabs}
\usepackage{subcaption}
\usepackage{hyperref}
\usepackage{tabularx}
\usepackage{url}
\usepackage[nohyperlinks, nolist]{acronym}
\usepackage{multirow}
\usepackage{tikz}
\usetikzlibrary{positioning, arrows, fit, backgrounds}
\usepackage{amsthm}
\usepackage[inkscapelatex=false]{svg}

\usepackage{xcolor}
\definecolor{TUMOrange}{RGB}{227, 114, 34}
\definecolor{TUMLightBlue}{RGB}{100, 160, 200}

\newtheorem{theorem}{Theorem}
\newtheorem{lemma}{Lemma}
\newtheorem{corollary}{Corollary}

\title{Safe Reinforcement Learning using Action Projection: \\ Safeguard the Policy or the Environment?}

\author{\name Hannah Markgraf$^{1, *}$ \quad 
      \name Shambhuraj Sawant$^3$ \quad 
      \name Hanna Krasowski$^2$ \quad 
      \name Lukas Sch{\"a}fer$^1$ \\[0.1cm]
      \name S\'ebastien Gros$^3$ \quad 
      \name Matthias Althoff$^1$  \\[0.3cm]
      \addr $^*$Corresponding author: \href{mailto:hannah.markgraf@tum.de}{hannah.markgraf@tum.de} \\[0.1cm]
      \addr $^1$Technical University of Munich, $^2$University of California, Berkeley, \\ $^3$Norwegian University of Science and Technology
      }

\def\month{03}  
\def\year{2026} 
\def\openreview{\url{https://openreview.net/forum?id=DDrGSEYxGU}} 

\begin{document}
\begin{acronym}
	\acro{rl}[RL]{reinforcement learning}
	\acro{mpc}[MPC]{model predictive control}
    \acro{mdp}[MDP]{Markov decision process}
    \acro{serl}[SE-RL]{safe environment RL}
    \acro{sprl}[SP-RL]{safe policy RL}
    \acro{pdf}[PDF]{probability density function}
    \acro{wrt}[w.r.t.]{with respect to}
    \acro{kkt}[KKT]{Karush-Kuhn-Tucker}
    \acro{crv}[CRV]{continuous random variable}
    \acro{a2c}[A2C]{Advantage Actor Critic}
    \acro{ppo}[PPO]{Proximal Policy Optimization}
    \acro{sac}[SAC]{Soft-Actor Critic}
    \acro{td3}[TD3]{Twin-delayed Deep Deterministic Policy Gradient}
    \acro{gae}[GAE]{generalized advantage estimation}
\end{acronym}

\newcommand{\pip}{\pi^\perp}

\maketitle

\begin{abstract}
Projection-based safety filters, which modify unsafe actions by mapping them to the closest safe alternative, are widely used to enforce safety constraints in reinforcement learning (RL). Two integration strategies are commonly considered: Safe environment RL (SE-RL), where the safeguard is treated as part of the environment, and safe policy RL (SP-RL), where it is embedded within the policy through differentiable optimization layers. Despite their practical relevance in safety-critical settings, a formal understanding of their differences is lacking.
In this work, we present a theoretical comparison of SE-RL and SP-RL. We identify a key distinction in how each approach is affected by action aliasing, a phenomenon in which multiple unsafe actions are projected to the same safe action, causing information loss in the policy gradients. In SE-RL, this effect is implicitly approximated by the critic, while in SP-RL, it manifests directly as rank-deficient Jacobians during backpropagation through the safeguard.
Our contributions are threefold: (i) a unified formalization of SE-RL and SP-RL in the context of actor-critic algorithms, (ii) a theoretical analysis of their respective policy gradient estimates, highlighting the role of action aliasing, and (iii) a comparative study of mitigation strategies, including a novel penalty-based improvement for SP-RL that aligns with established SE-RL practices. Empirical results support our theoretical predictions, showing that action aliasing is more detrimental for SP-RL than for SE-RL. However, with appropriate improvement strategies, SP-RL can match or outperform improved SE-RL across a range of environments. These findings provide actionable insights for choosing and refining projection-based safe RL methods based on task characteristics.  
\end{abstract}

\section{Introduction}
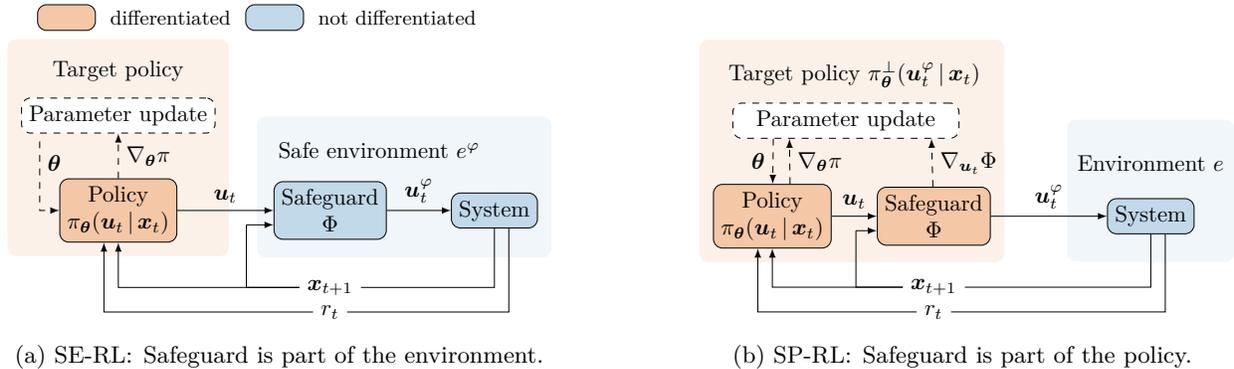
\begin{figure}[t]
    \centering
    \begin{subfigure}[b]{0.45\linewidth}
        \centering
        \resizebox{\linewidth}{!}{
        \begin{tikzpicture}[auto, node distance=2cm,>=latex]
        \node [draw, rounded corners, align=center, fill=TUMOrange!40] (policy) {Policy \\$\pi_{\boldsymbol\theta}(\vu_t \,|\, \vx_t)$};
        \node [draw, rounded corners, align=center, fill=TUMLightBlue!40, right=1.5cm of policy] (safeguard) {Safeguard \\$\Phi$};
        \node [draw, rounded corners, right=1.0cm of safeguard, fill=TUMLightBlue!40] (system) {System};
        \node[above=1.9cm of policy.center] (text) {Target policy};
        \node[above=1.2cm of policy.center, dashed, draw, rounded corners, fill=white] (opt) {Parameter update};
        \node[above left=0.7cm and 0.1cm of system.center] (text_env) {Safe environment $e^\varphi$};
        \begin{scope}[on background layer]
            \node[minimum height=2.2cm, fit={(policy) (text)  (opt)}, inner sep=6pt, rounded corners, fill=TUMOrange!10] (P2) {};
            \node[minimum height=2.2cm, inner sep=6pt, fit={(system) (safeguard) (text_env)}, fill=TUMLightBlue!10, rounded corners] (e1) {};
        \end{scope}
        
        \draw[->] (policy) -- node {$\vu_t$} (safeguard);
        \draw[->] (safeguard) -- node {$\vu_t^{\varphi}$} (system);
        \draw[->, dashed] (policy) -- node[right]{$\nabla_{\boldsymbol\theta} \pi$} (opt);
        \node[left=1.1cm of opt.south] (opt1) {};
        \draw[->, dashed] (opt1) |- node[above right= 0.5cm and 0.0cm]{${\boldsymbol\theta}$} (policy.west);
        
        \node[below left=0.1cm and 0.3cm of safeguard.west] (h1) {};
        \node[below=0.1cm of safeguard.west] (h2) {};
        \node[right=0.1cm of system.south] (h3) {};
        \node[left=0.1cm of policy.south] (h4) {};
        \node[below=0.5cm of safeguard] (output_x) {$\vx_{t+1}$};
        \node[below=0.9cm of safeguard] (output_r) {$r_t$};
        \draw[-] (system) |- (output_x);
        \draw[->] (output_x) -| (policy);
        \draw[-] (h3.center) |- (output_r);
        \draw[->] (output_r) -| (h4.center);
        \draw[-] (output_x) -| (h1.center);
        \draw[->] (h1.center) -- (h2.center);

        \node[draw, rounded corners, fill=TUMOrange!40, above left=2.8cm and 0.35 cm of policy.center, minimum width=0.9cm, minimum height=0.4cm] (legend1) {};
        \node[right=0.1cm of legend1] (l1) {\small differentiated};
        \node[draw, rounded corners, fill=TUMLightBlue!40, minimum width=0.9cm, minimum height=0.4cm, right=0.1cm of l1] (legend2) {};
        \node[right=0.1cm of legend2] (l2) {\small not differentiated};
        
    \end{tikzpicture}
    }
        \caption{\Ac{serl}: Safeguard is part of the environment.}
        \label{fig:p1}
    \end{subfigure}
    \hfill
    \begin{subfigure}[b]{0.45\linewidth}
        \centering
        \resizebox{\linewidth}{!}{
        \begin{tikzpicture}[auto, node distance=2cm,>=latex]
        \node [draw, rounded corners, align=center, fill=TUMOrange!40] (policy) {Policy \\$\pi_{\boldsymbol\theta} (\vu_t \,|\, \vx_t)$};
        \node [draw, rounded corners, align=center, right=0.7cm of policy, fill=TUMOrange!40] (safeguard) {Safeguard \\$\Phi$};
        \node[draw, dashed, rounded corners, align=center, fill=white, above right= 1.2cm and 0.3cm of policy.west, minimum width=3.5cm] (opt) {Parameter update};
        \node [draw, rounded corners, right=1.8cm of safeguard, fill=TUMLightBlue!40] (system) {System};
         \node[above left=2.2cm and 0.8cm of policy.center, anchor=west] (text) {Target policy $\pi_{\boldsymbol\theta}^\perp (\vu_t^\varphi \,|\, \vx_t)$};
        \node[above=0.3cm of system] (text_env) {Environment $e$};
        \begin{scope}[on background layer]
            \node[minimum height=2.2cm, fit={(policy) (safeguard) (text) (opt)}, inner sep=6pt, rounded corners, fill=TUMOrange!10] (P1) {};
            \node[minimum height=2.2cm, inner sep=1pt, fit={(system) (text_env)}, fill=TUMLightBlue!10, rounded corners] (e1) {};
        \end{scope}
        \draw[->] (policy) -- node {$\vu_t$} (safeguard);
        \node[left=0.75cm of opt.south] (opt1) {};
        \node[right=1.2cm of opt.south] (opt2){};
        \node[left=1.0cm of opt.south] (opt3) {};
        \node[right=0.15cm of policy.north] (pol1) {};
        \draw[->, dashed] (pol1.center) -- node[right] {$\nabla_{\boldsymbol\theta} \pi$} (opt1.center);
        \draw[->, dashed] (safeguard) -- node[right] {$\nabla_{\vu_t}\Phi$} (opt2.center);
        \draw[->, dashed] (opt3.center) -- node[left] {${\boldsymbol\theta}$} (policy);
        \draw[->] (safeguard) -- node {$\vu_t^{\varphi}$} (system);
        
        \node[below=0.4cm of safeguard] (output_x) {$\vx_{t+1}$};
        \node[below=0.8cm of safeguard] (output_r) {$r_t$};
        \node[below left=0.1cm and 0.2cm of safeguard.west] (h1) {};
        \node[below=0.1cm of safeguard.west] (h2) {};
        \node[right=0.1cm of system.south] (h3) {};
        \node[left=0.1cm of policy.south] (h4) {};
        \draw[-] (system) |- (output_x);
        \draw[->] (output_x) -| (policy);
        \draw[-] (output_x) -| (h1.center);
        \draw[->] (h1.center) -- (h2.center);
        \draw[-] (h3.center) |- (output_r);
        \draw[->] (output_r) -| (h4.center);
            
        \node[draw=white, rounded corners, fill=white, above left=2.8cm and 0.35 cm of policy.center, minimum width=0.9cm, minimum height=0.4cm] (legend1) {};
    \end{tikzpicture}
        }
        \caption{\Ac{sprl}: Safeguard is part of the policy.}
        \label{fig:p2}
    \end{subfigure}
    \caption{In provably safe reinforcement learning, we can either consider the safeguarding as part of the environment (figure \ref{fig:p1}) or as part of the policy (figure \ref{fig:p2}).}
    \label{fig:perspectives}
\end{figure}

For safety-critical environments, \ac{rl} policies have to be verified or safeguarded to ensure safety specifications at all times. Provably safe RL through closest-point projection \citep{krasowski2023provably}, also often called safety filtering, is a widely used approach that provides safety guarantees during both training and deployment. The projection operation adjusts unsafe actions to the closest safe action by solving an optimization problem. This operation is usually differentiable \citep{gros2020safe}, allowing for two different formulations: 
\begin{itemize}
    \item[(a)] \Ac{serl}: The projection is treated as part of the unknown environment dynamics, requiring the critic to understand its effect indirectly. Intuitively, the agent learns an unsafe policy that acts in a safeguarded environment \citep{hunt2021verifiably}.
    \item [(b)] \Ac{sprl}: The projection is integrated into the policy itself, meaning that gradients are backpropagated through the safeguard. In \ac{sprl}, the objective is to approximate the optimal safeguarded policy for the original unsafe environment \citep{pham2018optlayer}.
\end{itemize}
Figure~\ref{fig:perspectives} illustrates the structural differences between \ac{serl} and \ac{sprl}. A key advantage of \ac{serl} is that it leaves the underlying \ac{rl} algorithm unchanged, preserving any existing theoretical guarantees \citep{hunt2021verifiably}. On the other hand, \ac{sprl} promises a more direct informing of the agent about the impact of the safeguarding by including the sensitivity of the safe action with respect to the unsafe action in the policy gradient estimator. However, \ac{sprl} requires embedding the safeguard within the policy, typically using differentiable optimization layers \citep{agrawal2019differentiable}, and incurs additional computational cost due to sensitivity analysis. These trade-offs lead to a central question: should one safeguard the environment or the policy? 

Although empirical comparisons exist \citep{pham2018optlayer, kasaura2023benchmarking}, a theoretical foundation for understanding the trade-offs between \ac{serl} and \ac{sprl} is still lacking. In this work, we close this gap by developing a formal framework that clarifies their relationship, identifying both equivalences and key differences. As discussed later in section \ref{sec:action_aliasing}, a central difference lies in how both approaches are affected by what we refer to as \textit{action aliasing}: In closest-point projection, multiple unsafe actions are mapped to the same safe action, resulting in identical returns. 
Action aliasing has been well studied in the context of \ac{sprl}. It has been shown to lead to a rank-deficient Jacobian of the safeguard \citep{gros2020safe}, also referred to as the zero-gradient problem \citep{lin2021escaping}. The zero-gradient problem has been associated with degraded performance \citep{pham2018optlayer, bhatia2019resource} and is commonly addressed through alternative loss functions \citep{bhatia2019resource, chen2021enforcing} or modified policy update rules \citep{pham2018optlayer}.
Similarly, performance issues linked to action aliasing have been reported for \ac{serl} \citep{krasowski2023provably}, and are often mitigated by introducing penalty terms in the reward function that penalize the distance between the original and projected actions \citep{wabersich2021predictive, wang2022ensuring, markgraf2023safe, bejarano2024safety}. However, a formal comparison of the action aliasing effect on learning in \ac{serl} versus \ac{sprl}, and of the differences in improvement strategies, does not yet exist.

Our contributions are as follows:

\begin{itemize}
\item We develop a unified formalization of \ac{serl} and \ac{sprl} in terms of the underlying \ac{mdp}, value functions, and policy gradient estimators;

\item We prove that the optimal value functions of \ac{serl} and \ac{sprl} coincide under mild assumptions;

\item We formalize the impact of action aliasing on \ac{serl}, highlighting how it differs from the known zero-gradient problem induced by action aliasing in \ac{sprl};

\item We identify fundamental differences in how action aliasing is mitigated in \ac{serl} and \ac{sprl}, and propose a novel remedy for \ac{sprl} that aligns more closely with the penalty-based strategy used in \ac{serl};

\item We perform an empirical comparison of \ac{serl} and \ac{sprl}, emphasizing the importance of adaptation strategies for handling action aliasing in both deterministic and stochastic policy settings.
\end{itemize}

The remainder of this study is structured as follows: In section \ref{sec:related_work}, we provide an overview of the related literature. The theoretical background for \ac{serl} and \ac{sprl} is established in section \ref{sec:prelim}. We then formalize the problem statement in section \ref{sec:problem}. Sections \ref{sec:main} and \ref{sec:comparison} provide a formal definition and a comparative analysis of the two approaches, respectively. We discuss mitigation strategies for action aliasing and suggest a new alternative for \ac{sprl} in section \ref{sec:improvs}, before conducting a thorough experimental evaluation in section \ref{sec:experiments}. We discuss our findings in section \ref{sec:discussion} before concluding in section \ref{sec:conclusion}. 

\section{Related Work}\label{sec:related_work}
Safe \ac{rl} augments \ac{rl} algorithms with mechanisms to increase the probability of learning and deploying safe policies, or to ensure hard safety guarantees for the policy \citep{Garcia2015}. Specifically, provably safe \ac{rl} encompasses approaches for which hard safety guarantees are provided during training and deployment \citep{krasowski2023provably}. Provably safe \ac{rl} can be further categorized by means of ensuring that only safe actions are executed. The first category relies on pre-characterizing the set of safe actions through specific structural properties, as it is often necessary to compute center or boundary points efficiently. It comprises methods such as mapping unsafe actions to the interior of the safe action set \citep{tabas2022computationally, kasaura2023benchmarking} or sampling from within it \citep{stolz2024excluding}. The second category of safeguarding mechanisms works directly with the safety constraints themselves, defined, for example, through control barrier functions \citep{marvi2022reinforcement, wang2022ensuring} or predictive filters \citep{selim2022safe, Wabersich2023, markgraf2023safe, gros2020safe}. Within this group, closest-point projection is particularly prevalent in continuous control settings, mapping unsafe actions to their nearest safe counterpart by solving a constrained optimization problem. 

While effective and widely used, action projection introduces a key limitation that we term action aliasing: All unsafe policy actions that lie within the normal cone to the boundary of the safe action set are projected to the same safe action (see lemma \ref{def:action_aliasing}). Consequently, they all incur the same reward, no matter how close the policy action was to the safe action. In the context of \ac{serl}, action aliasing has not been theoretically analyzed, although it has been empirically acknowledged by some studies \citep{wang2022ensuring, krasowski2023provably, markgraf2023safe}. In contrast, in \ac{sprl}, the impact of action aliasing is well understood. Here, the projection is commonly integrated into the policy using differentiable optimization layers \citep{agrawal2019differentiable} as the last layer of the policy network to retain gradient flow through the safeguard \citep{pham2018optlayer, dalal2018safe, bhatia2019resource, chen2021enforcing, kasaura2023benchmarking}. Consequently, action aliasing directly affects the policy gradient computation, eliminating components in the normal direction of the mapping \citep{gros2020safe, walter2025leveraging}. 

The approaches for addressing action aliasing differ in \ac{serl} and \ac{sprl}. A common remedy in \ac{serl} is to augment the reward with a penalty proportional to the action adjustment \citep{wabersich2021predictive, markgraf2023safe, stanojev2023safe, bejarano2024safety, kasaura2023benchmarking, dawood2025safe}. As shown in \citet{markgraf2023safe}, agents trained with proportional penalties often outperform those trained with constant or no penalties. Recently, \citet{bejarano2024safety} confirmed this observation in quadrotor hardware experiments.
In \ac{sprl}, additional policy loss terms are often used. For example, \citet{bhatia2019resource} employs a loss term that is proportional to the safety constraint violation. Similarly, \citet{chen2021enforcing} proposes a loss term that is proportional to the Euclidean distance between the unsafe and the corresponding safe action. \citet{pham2018optlayer} proposes a two-step gradient step approach in which a first update step is calculated for the unsafe action (with a penalty, following \ac{serl}), followed by an update step on the safe action. 

Most existing studies treat projection safeguarding separately in the context of either \ac{serl} or \ac{sprl}, leaving a theoretical comparison of the two approaches unexplored. \citet{kasaura2023benchmarking} provides a valuable empirical comparison of both basic and improved versions of these methods, focusing on deterministic policies with static safety constraints. Building on their empirical insights, our work contributes a comprehensive theoretical framework that explains the fundamental differences between \ac{serl} and \ac{sprl}, including their respective improvement strategies for handling action aliasing. We extend the empirical analysis to include both deterministic and stochastic policies, consider benchmark problems with state-dependent safety constraints, and focus on appropriately scaling penalties in \ac{serl} and additional loss terms in \ac{sprl}.

\section{Preliminaries}\label{sec:prelim}
Policy-based \ac{rl} algorithms augmented with safety measures have proven successful in safety-critical tasks \citep{krasowski2023provably}. Therefore, we provide an overview of the basic concepts of these algorithms before detailing how to differentiate between \ac{serl} and \ac{sprl} for projection-based safeguarding in section~\ref{sec:main}.

\subsection{Reinforcement Learning}\label{sec:RL}

An \ac{mdp} is a tuple $\left({\sX}, \sU, p_r, p_x, \gamma \right)$, where $\sX$ and $\sU$ are the state space and action space. We assume $\rvx_t$, $\rvu_t$, and $\rr_t$ to be \acp{crv} modeling the state, action, and reward at time step $t$, respectively. The transition function $p_x: \sX \times \sU \times \sX \rightarrow \sR_{\geq0}$ denotes the probability density $p_x(\vx_{t+1} \,|\, \vx_t, \vu_t)$ of transitioning to a state $\vx_{t+1}$ when applying the action $\vu_t$ in state $\vx_t$ \citep[section 1.1]{van2012reinforcement}.
The reward for a transition is defined through the probability density $p_r(r_t \, |\, \vx_t, \vu_t)$, where $p_r: \sX \times \sU \times \sR \rightarrow \sR_{\geq 0}$ \citep[section~1.1]{van2012reinforcement}. 
The discount factor $0 \leq \gamma \leq 1$ is used to discount long-term rewards and serves as a simplistic model of the probabilistic lifetime of the MDP. 

We define the discounted return at time $t$ as \citep[equation~3.8]{sutton2018reinforcement}
\begin{equation}\label{eq:return}
    \rg_t = \sum_{k=0}^\infty \gamma^k \rr_{t+k+1}.
\end{equation}
The goal of \ac{rl} is to find the policy $\pi$ associated with a probability density function $\pi(\vu_t \, | \, \vx_t)$ over actions given state $\vx_t$, that maximizes the expected return \citep[equation~13.4]{sutton2018reinforcement}
\begin{equation}\label{eq:rl_objective}
    J(\pi) = \E_{\pi} \left[\rg_0 \,|\, \vu_t \sim \pi(\cdot \,|\, \vx_t) \right],
\end{equation}
where $\E_\pi[\cdot]$ refers to the expected value of a random variable given that the agent follows the policy $\pi$. 
Note that in the deterministic policy case, states are directly mapped to actions, such that $\vu_t = \pi(\vx_t)$. 

To find the optimal policy $\pi^* = \argmax_{\pi \in \Pi} J(\pi)$ \citep[equation~2]{gros2020safe}, most \ac{rl} algorithms use various forms of value function estimation. A state value function $v_\pi(\vx_t)~=~\E_{\pi} \left[\rg_t \,|\, \rvx_t = \vx_t \right]$ \citep[equation 3.12]{sutton2018reinforcement} expresses how good it is to be in a certain state $\vx_t$ provided that policy $\pi$ is in use. A state-action value function $q_\pi(\vx_t, \vu_t)~=~\E_{\pi} \left[\rg_t \,|\, \rvx_t = \vx_t, \rvu_t = \vu_t\right]$\citep[equation 3.13]{sutton2018reinforcement} expresses how good it is to use action $\vu_t$ in state $\vx_t$, provided that policy $\pi$ is in use after that.
Finally, the advantage of taking an action $\vu_t$ in state $\vx_t$ is commonly defined as $a_\pi(\vx_t, \vu_t) = q_\pi(\vx_t, \vu_t) - v_\pi(\vx_t)$.
All optimal policies $\pi^*$
satisfy the same so-called optimal value functions  
\begin{align}
    v^*(\vx_t) &= v_{\pi^*}(\vx_t) \quad \forall \vx_t \in \sX, \\
    q^*(\vx_t, \vu_t) &= q_{\pi^*}(\vx_t, \vu_t) \quad \forall \vx_t\in \sX, \vu_t \in \sU.
\end{align} 

\subsection{Policy Gradient Algorithms}
The policy $\pi$ can be inferred directly from the value function, a central concept of so-called value-based RL algorithms. However, these algorithms are mainly designed for discrete action spaces. In continuous and high-dimensional action domains, it is more common to directly learn a policy $\pi_{\boldsymbol\theta}$ parameterized by ${\boldsymbol\theta}$, also referred to as policy-based RL. Most modern algorithms use actor-critic algorithms, combining concepts from value-based and policy-based RL by simultaneously learning a policy (the actor) and a value function (the critic) parameterized by $\boldsymbol\phi$. 
To find the optimal policy parameters $\boldsymbol\theta^*$, actor-critic algorithms perform repeated updates using ${\boldsymbol\theta} \leftarrow {\boldsymbol\theta} + \alpha \nabla_{\boldsymbol\theta} J(\pi_{\boldsymbol\theta})$ \citep[equation~13]{van2012reinforcement}
where $\alpha$ is a learning rate. The gradient $\nabla_{\boldsymbol\theta} J(\pi_{\boldsymbol\theta})$ cannot be computed analytically since the reward function and the transition probability distribution are not known explicitly. RL algorithms thus estimate the gradient from data, more specifically from tuples $(\vx_t, \vu_t, \vx_{t+1}, r_t)$. According to the policy gradient theorem, an unbiased estimate of $\nabla_{\boldsymbol\theta} J(\pi_{\boldsymbol\theta})$ is given by \citep[theorem 1]{silver2014deterministic}
\begin{equation} \label{eq:policy_gradient_estimate_determ}
    \nabla_{\boldsymbol\theta} J(\pi_{\boldsymbol\theta}) = \E_\pi \left[\nabla_{\boldsymbol\theta}  \pi_{\boldsymbol\theta}(\vx_t) \nabla_{\vu_t} q_{\pi, \boldsymbol\phi}(\vx_t, \vu_t) \right]
\end{equation}
for deterministic policies and a parameterized critic $q_{\pi,\boldsymbol\phi}$, and \citep[equation~15]{van2012reinforcement}
\begin{equation} \label{eq:policy_gradient_estimate}
    \nabla_{\boldsymbol\theta} J(\pi_{\boldsymbol\theta}) = \E_\pi \left[\Psi(\vx_t,\vu_t) \nabla_{\boldsymbol\theta} \log \pi_{\boldsymbol\theta}(\vu_t\,|\,\vx_t) \right]
\end{equation}
for stochastic policies, where $\Psi$ differs depending on the algorithm. Possible choices include $q_{\pi,\boldsymbol\phi}(\vx_t, \vu_t); a_{\pi, \boldsymbol\phi}(\vx_t, \vu_t)$\citep{schulman2015high}, which vary in the variance of the gradient estimate.

Actor-critic algorithms with stochastic policies such as \ac{a2c} \citep{mnih2016asynchronous} or \ac{ppo} \citep{schulman2017proximal} that learn a state value $v_{\pi, \boldsymbol\phi}$ use $\Psi~=~\hat{a}(\vx_t,\vu_t)$, where $\hat{a}$ is an estimate of the true advantage function.
The most common estimator is the \ac{gae} \citep[equation~16]{schulman2015high}
\begin{equation}\label{eq:gae}
    \hat{a}_t^{\text{GAE}} = \sum_{l=0}^\infty (\gamma \lambda)^l \delta_{t+l}^v,
\end{equation}
where $\lambda$ is a hyperparameter and $\delta_t^v$ is the temporal difference residual \citep[equation~10]{schulman2015high}
\begin{equation}\label{eq:td_residual_v}
    \delta_t^v = r_t + \gamma v_{\pi,\boldsymbol\phi}(\vx_{t+1}) - v_{\pi,\boldsymbol\phi}(\vx_t).
\end{equation}

To find the optimal parameters $\boldsymbol\phi^*$ of the value function $q_{\pi,\boldsymbol\phi}$, actor-critic algorithms minimize the expected squared error between the current value function estimate and a target $y$ using \citep[equation~4]{lillicrap2015continuous}
\begin{equation}\label{eq:critic_loss}
    L_{\boldsymbol\phi} = \E \bigg[\underbrace{(r_t + \gamma q_{\pi, \tilde{\boldsymbol\phi}}(\vx_{t+1},\vu_{t+1})}_y - q_{\pi, \boldsymbol\phi}(\vx_t, \vu_t))^2\bigg],
\end{equation}
or the equivalent equation for a state value function $v_{\pi, \boldsymbol\phi}$. Here, $q_{\pi,\tilde{\boldsymbol\phi}}$ is a target critic that is slowly updated using Polyak averaging. Alternative choices for the target include using the return $\rg_t$ directly, or more advanced formulations that address overestimation bias using the minimum value of two target critic networks \citep{fujimoto2018addressing}.

\section{Problem Statement} \label{sec:problem}
We consider a system with dynamics defined through the transition function $p_x$ that is subject to state-dependent safety constraints of the form
\begin{equation}
    \label{eq:safety_specs}
    g(\vx_t, \vu_t) \leq \mathbf{0} \,,
\end{equation}
which must be satisfied at all times.
To enforce these constraints, we define a state-dependent safe action set $\sU^\varphi_{\vx_t} \subseteq \sU$, such that for any $\vu_t \in \sU^\varphi_{\vx_t}$, there exists a policy that, when applied consecutively from time $t+1$ onward, ensures satisfaction of \eqref{eq:safety_specs} at all times $t' \geq t$. We assume that $\sU^\varphi_{\vx_t}$ can be represented as 
\begin{equation}
    \label{eq:implicit_safe_set}
    \sU^\varphi_{\vx_t} = \{\vu_t \,|\, s(\vx_t, \vu_t) \leq \mathbf{0} \},
\end{equation}
where constraints $s$ could, for example, be formulated using control barrier functions, predictive filters, or, as in our experiments, robust control invariant (RCI) sets (see appendix \ref{app:zonotopes}). Furthermore, we define $\tilde{\sX} \subseteq \sX$ as the set of states for which $\sU^\varphi_{\vx_t}$ is non-empty.

Given the safe action set, we introduce a safeguard $\Phi: \tilde{\sX} \times \sU \rightarrow \sU^\varphi$ into the interaction between the \ac{rl} policy and the system that ensures that unsafe inputs are projected to the constraint boundary by solving 
\begin{subequations}\label{eq:projection_problem}
\begin{align}
\label{eq:projection}
    \Phi(\vx_t, \vu_t)= \argmin_{\tilde{\vu}_t \in \sU}&\quad \frac{1}{2} \|\tilde{\vu}_t - \vu_t \|^2_2    \\
    \text{s.t.} \quad & s(\vx_t,\tilde{\vu}_t) \leq 0\label{eq:projection_constraints},  
\end{align}
\end{subequations}
where $\vu_t \sim \pi(\cdot \,|\, \vx_t)$. As a result, only safe actions $\vu^\varphi_t = \Phi(\vx_t, \vu_t)$ can be applied to the system, and the return in \eqref{eq:return} depends on a sequence of safe actions and states. Note that the mapping in \eqref{eq:projection_problem} only modifies unsafe actions. To ensure that $\Phi$ is a deterministic operator, we assume that $\sU^\varphi_{\vx_t}$ is convex. This is a mild assumption since safe sets are commonly constructed as convex inner approximations in practice due to their simple parametrization (e.g., polytopes, ellipsoids), computational tractability, and the uniqueness of closest-point projections.

As shown in figure \ref{fig:perspectives}, we can use two different architectures to integrate $\Phi$ into the \ac{rl} training loop to maximize the expected return in \eqref{eq:rl_objective}: We can consider the safeguard to be part of the black-box environment dynamics and aim to approximate the optimal unsafe policy $\pi^* \in \Pi$ for the safeguarded environment. Or we can learn a safe policy $\pip \in \Pi^\perp$, where $\Pi^\perp \subseteq \Pi$ is the set of safe policies. Note that for any safe policy, the probability density function $\pip(\vu^\varphi_t \,|\, \vx_t)$ is zero for all unsafe actions. 
In this work, we investigate whether one solution approach dominates the other, both theoretically and empirically. 
\paragraph{Notation:} In the remainder of this work, we will omit the subscript $t$ when possible and use $\vx, \vx',\vu,r$ instead of $\vx_t, \vx_{t+1},\vu_t, r_t$ for brevity. Furthermore, we will omit the subscripts ${\boldsymbol\theta}$ and $\boldsymbol\phi$ to improve readability. 

\section{Perspectives on Safe Reinforcement Learning}\label{sec:main}

We develop a unified theoretical framework for both approaches to safe \ac{rl} with action projection by specifying the corresponding value functions and policy gradient estimates. An overview is provided in table \ref{tab:overview_equations}. For brevity, we limit our analysis to \ac{rl} algorithms derived directly from the policy gradient theorem, as including methods with additional modifications (e.g., \ac{ppo}, \ac{sac}) would require extensive theoretical derivations beyond the scope of this paper. Similar to the work in \citet{gros2020safe}, we distinguish between stochastic and deterministic policies. We then discuss the equivalences and differences of both approaches in section \ref{sec:comparison}.

\subsection{Safe Environment Reinforcement Learning (SE-RL)}\label{sec:serl}
In \ac{serl}, the safeguard is considered part of the system dynamics that are unknown to the \ac{rl} agent. This changes the \ac{mdp} from section \ref{sec:RL} to $M^\text{SE} =(\tilde{\sX},\sU, p_r^\text{SE}, p_x^\text{SE}, \gamma)$, with $p_x^\text{SE}: \tilde{\sX} \times \sU \times \tilde{\sX} \rightarrow \sR_{\geq 0}$ and $p_r^\text{SE}: \tilde{\sX} \times \sU  \times \sR \rightarrow \sR_{\geq 0}$. Please note that the support of these densities changes compared to the original MDP introduced in section \ref{sec:RL}. Here, $p_x^\text{SE}(\vx' \, | \, \vx, \Phi(\vx,\vu))$ and $p_r^\text{SE}(r \, | \,\vx, \Phi(\vx,\vu))$ are defined as the composition of the safeguard mapping $\Phi$ with the original transition functions introduced in section \ref{sec:RL}.
Accordingly, the environment receives a (potentially) unsafe action $\vu \sim \pi$ that is projected internally. Consequently, the value functions are learned for the unsafe policy such that
\begin{subequations}
\begin{align}
   q_\pi^{\text{SE}}(\vx, \vu) &= \E_\pi\left[\rg_t \,|\, \rvx_t=\vx, \rvu_t=\vu\right], \label{eq:q1_bellman} \\
   v_\pi^{\text{SE}}(\vx) &= \E_\pi\left[\rg_t \,|\, \rvx_t=\vx\right]. \label{eq:v1_bellman}
\end{align}
\end{subequations}

The policy gradient estimate is computed as given in \eqref{eq:policy_gradient_estimate_determ} and \eqref{eq:policy_gradient_estimate}. Note that the critic forming this estimate is not aware of the safeguard and has to assess its impact through the data alone.

\subsection{Safe Policy Reinforcement Learning (SP-RL)}
\begin{table*}[]
    \centering
    \caption{Overview of the \ac{mdp} formulations and learning equations for \ac{serl} and \ac{sprl} with \textcolor{TUMLightBlue}{stochastic} and \textcolor{TUMOrange}{deterministic} policies, respectively.}
    \label{tab:overview_equations}
    \renewcommand{\arraystretch}{1.5}
    \begin{tabularx}{\textwidth}{X l l}
    \toprule
         & \textbf{\ac{serl}} & \textbf{\ac{sprl}} \\ \midrule
         \multirow{3}{8em}{MDP} & $M^\text{SE} =(\tilde{\sX},\sU, p_r^\text{SE}, p_x^\text{SE}, \gamma)$ & $M^\text{SP} =(\tilde{\sX},\sP(\sU^\varphi), p_r^\text{SP}, p_x^\text{SP}, \gamma)$ \\ 
         & $p_x^\text{SE}: \tilde{\sX} \times \sU \times \tilde{\sX} \rightarrow \sR_{\geq 0}$ & $p_x^\text{SP}: \tilde{\sX} \times \sP(\sU^\varphi) \times \tilde{\sX} \rightarrow \sR_{\geq 0}$\\ 
         & $p_r^\text{SE}: \tilde{\sX} \times \sU \times \sR \rightarrow \sR_{\geq 0}$ & $p_r^\text{SP}: \tilde{\sX} \times \sP(\sU^\varphi) \times \sR \rightarrow \sR_{\geq 0}$\\ \midrule
        Policy & $\textcolor{TUMLightBlue}{\pi: \tilde{\sX} \times \sU \rightarrow \sR_{\geq 0}}$; $\textcolor{TUMOrange}{\pi: \tilde{\sX} \rightarrow \sU}$ & $\textcolor{TUMLightBlue}{\pip: \tilde{\sX} \times \sP(\sU^\varphi) \rightarrow \sR_{\geq 0}}$; $\textcolor{TUMOrange}{\pip: \tilde{\sX} \rightarrow \sP(\sU^\varphi)}$ \\
         
         \midrule
         \multirow{2}{8em}{Value \\functions} & $q_\pi^{\text{SE}}(\vx, \vu) =  \E_\pi\left[\rg_t \,|\, \rvx_t=\vx, \rvu_t=\vu\right]$ & $q_{\pip}^{\text{SP}}(\vx, \vu^\varphi) = \E_{\pip} \left[\rg_t \,|\, \rvx_t=\vx, \rvu_t^\varphi=\vu^\varphi\right]$\\ 
         & $v_\pi^{\text{SE}}(\vx) = \E_\pi \left[\rg_t \,|\, \rvx_t=\vx \right]$ & $v_{\pip}^{\text{SP}}(\vx) = \E_{\pip}\left[\rg_t \,|\, \rvx_t=\vx\right]$\\ \midrule
         \multirow{2}{8em}{Policy \\gradient \\estimates} & $\textcolor{TUMLightBlue}{\nabla_{\boldsymbol\theta} J(\pi) = \E_\pi \left[\Psi^\text{SE}_\pi(\vx,\vu) \nabla_{\boldsymbol\theta} \log \pi(\vu\,|\,\vx)\right]}$ & $\textcolor{TUMLightBlue}{\nabla_{\boldsymbol\theta} J(\pip) = \E_{\pip} \left[\Psi^\text{SP}_{\pip}(\vx,\vu^\varphi) \nabla_{\boldsymbol\theta} \log \pi(\vu \,|\, \vx) \right]}$\\ 
         & $\textcolor{TUMOrange}{\nabla_{\boldsymbol\theta} J(\pi) = \E_\pi \left[\nabla_{\boldsymbol\theta} \pi(\vx) \nabla_\vu q^\text{SE}_\pi(\vx, \vu) \right]}$ &  $\textcolor{TUMOrange}{\nabla_{\boldsymbol\theta} J(\pip) = \E_{\pip} \left[\nabla_{\boldsymbol\theta} \pip(\vx) \nabla_{\vu^\varphi} q^\text{SP}_{\pip}(\vx, \vu^\varphi) \right]}$\\ 
         \bottomrule
    \end{tabularx}
\end{table*}
If we consider the safeguard to be a part of the policy, sampled actions are adjusted by a differentiable optimization layer, such that the final policy action is a random variable $\rvu^\varphi_t := \Phi(\rvx_t, \rvu_t)$.  The transition function and the probability density of the reward of the \ac{mdp} $M^\text{SP}$ for \ac{sprl} are then defined as $p_x^\text{SP}: \tilde{\sX} \times \sP(\sU^\varphi) \times \tilde{\sX} \rightarrow \sR_{\geq 0}$ and $p_r^\text{SP}: \tilde{\sX} \times \sP(\sU^\varphi) \times \sR \rightarrow \sR_{\geq 0}$, respectively, 
where the action space $\sP(\sU^\varphi)$ is the power set of all $\sU^\varphi_\vx$. 
We aim to learn a value function for the safe actions such that
\begin{subequations}
\begin{align}
     q_{\pip}^{\text{SP}}(\vx, \vu^\varphi) &= \E_{\pip} \left[\rg_t \,|\, \rvx_t=\vx, \rvu_t^\varphi=\vu^\varphi\right], \\
     v_{\pip}^{\text{SP}}(\vx) &= \E_{\pip} \left[\rg_t \,|\, \rvx_t=\vx \right]. \label{eq:v2_bellman}
\end{align}
\end{subequations}

Note that this requires us to also project the action $\vu'$ when computing the state-action value $q_{\pi, \tilde{\Phi}}(\vx', \vu^{\varphi'})$ for the target $y$ given in \eqref{eq:critic_loss}.
To obtain the policy gradient estimates, we must distinguish between deterministic and stochastic policies. 

\subsubsection{Deterministic Policies}\label{subsec:det_policies}
The deterministic policy gradient estimate in \ac{sprl} is defined by 
\begin{equation}
    \nabla_{\boldsymbol\theta} J(\pip) = \E_{\pip} \left[\nabla_{\boldsymbol\theta} \pip(\vx) \nabla_{\vu^\varphi} q^\text{SP}_{\pip}(\vx, \vu^\varphi) \right].
\end{equation}
Computing the gradient $\nabla_{\boldsymbol\theta} J(\pip)$ requires differentiating through the safeguard $\Phi$.
We can apply the chain rule to obtain
\begin{equation} \label{eq:grad_p2_deterministic}
     \nabla_{\boldsymbol\theta} \pip(\vx) = \nabla_{\vu} \Phi(\vx,\vu) \nabla_{\boldsymbol\theta} \pi(\vx).
\end{equation}
The sensitivity of the safeguard $\nabla_{\vu} \Phi(\vx,\vu)$ can be obtained using the implicit function theorem as described in appendix \ref{app:diff_safeguard}.

\subsubsection{Stochastic Policies}\label{sec:sprl_stochastic_policies}
To find the optimal parameters of the safeguarded policy $\pip$ in the stochastic policy case, we use the policy gradient estimate
\begin{equation}\label{eq:pg_estimate_SPRL}
    \nabla_{\boldsymbol\theta} J(\pip) = \E_{\pip} \left[\Psi^\text{SP}_{\pip}(\vx,\vu^\varphi) \nabla_{\boldsymbol\theta} \log \pip(\vu^\varphi \,|\, \vx) \right],
\end{equation}
which depends on the probabilities of safe actions given states as well as on the value function estimate for the safe policy $\Psi^\text{SP}_{\pip}$. While the unsafe policy $\pi(\vu\,|\,\vx)$ features a bounded probability density, the safe policy $\pip(\vu^\varphi\,|\,\vx)$ takes on a Dirac-like structure on the boundary of the safe set as illustrated in \citet[figure 2]{gros2020safe}. In general, no closed-form expression of $\pip(\vu^\varphi\,|\,\vx)$ exists, making it difficult to provide a gradient estimate for a projected stochastic policy. However, in \citet[Proposition 2]{gros2020safe}, the authors show that we obtain an unbiased gradient estimate using 
\begin{equation}\label{eq:grad_estimate_sprl_stochastic}
    \nabla_{\boldsymbol\theta} J(\pip) = \E_{\pip}\left[\Psi^\text{SP}_{\pip}(\vx, \vu^\varphi) \nabla_{\boldsymbol\theta} \log \pi(\vu\,|\,\vx)\right].\footnote{An alternative for \ac{sprl} with stochastic policies is proposed by \citet{chen2021enforcing}. Given an unsafe policy represented by a Gaussian distribution $\mathcal
N(\vmu_{\boldsymbol\theta}, \mSigma_{\boldsymbol\theta})$, they first project the mean using $\vmu_{\boldsymbol\theta}^\perp(\vx) = \Phi(\vmu_{\boldsymbol\theta}, \vx)$ and then sample $\vu^s \sim \mathcal{N}(\vmu_{\boldsymbol\theta}^\perp, \mSigma_{\boldsymbol\theta})$. However, since this action is not guaranteed to satisfy the constraints, they project again to obtain $\vu^\varphi = \Phi(\vu^s, \vx)$. 
Since this approach does not solve the problem of finding a closed-form solution for $\pip(\vu^\varphi\,|\,\vx)$, it is not further considered here.}
\end{equation}

\section{Comparative Analysis of SE-RL and SP-RL}\label{sec:comparison}
Having established \ac{serl} and \ac{sprl} as two distinct solution paradigms for the same task, we now examine their theoretical relationship and practical differences. Since the learning equations differ between approaches, identical initial policy and critic parameters will yield different updates, raising a fundamental question: do both frameworks converge to equivalent (sub-)optimal solutions?

\subsection{Theoretical Equivalence of Optimal Solutions}\label{sec:eq_serl_sprl}
We first establish that, in principle, both approaches target the same optimal value function.
\begin{theorem}\label{theorem:eq_opt_val_functions}
    For any given task, let $v_\pi^\text{SE}$ be a value function for policy $\pi$ interacting with the \ac{mdp} $M^\text{SE}$ and $v_{\pip}^\text{SP}$ a value function for policy $\pip$ interacting with $M^\text{SP}$. Then, a value function $v^{\text{SE}^*}$ that is optimal for $M^\text{SE}$ is also an optimal value function $v^{\text{SP}^*}$ for $M^\text{SP}$.
\end{theorem}
The proof is provided in appendix \ref{app:proof_theorem_opt_value_functions}. Theorem 1 states that, given a specific task, a policy $\pi^*$ that is optimal in the \ac{serl} framework yields the same expected return as an optimal policy ${\pip}^*$ in the \ac{sprl} framework.
While this theoretical result provides important insight, it offers limited practical guidance for continuous state and action spaces where optimal policies are rarely found \citep{van2012reinforcement}. Subsequently, we derive a more practical result for a certain group of algorithms. 

\subsection{Practical Equivalence for Stochastic Policies}
Algorithms employing stochastic policies and \ac{gae} (e.g., \ac{a2c}) learn a state value function $v_\pi(\vx)$ to compute $\hat{a}^\text{GAE}$. The following lemma shows that in this case, the differences between \ac{serl} and \ac{sprl} disappear.
\begin{lemma}\label{lemma:serl_sprl_stochastic}
    Let $\pi_{\boldsymbol\theta}, \pip_{\boldsymbol\theta}$ and $v^\text{SE}_{\pi,\boldsymbol\phi}, v^\text{SP}_{\pip,\boldsymbol\phi}$ denote the parameterized policies and value functions for a given task when employing \ac{serl} or \ac{sprl}, respectively. Define the advantage estimates used in the policy gradient estimates in \ac{serl} (\eqref{eq:policy_gradient_estimate}) and \ac{sprl} (\eqref{eq:grad_estimate_sprl_stochastic}) as 
    \begin{equation*}
        \Psi^\text{SE}_{\pi}(\vx,\vu)=\hat{a}^\text{GAE}(\vx, \vu) \quad \text{and} \quad  \Psi^\text{SP}_{\pip}(\vx,\vu^\varphi)=\hat{a}^\text{GAE}(\vx, \vu^\varphi),
    \end{equation*}
    respectively, 
    where $\hat{a}^\text{GAE}$ is computed using \eqref{eq:gae}. Then, for any initial parameters ${\boldsymbol\theta}_0$ and $\boldsymbol\phi_0$, the parameter updates ${\boldsymbol\theta}_{k+1}~\leftarrow~{\boldsymbol\theta}_k~+~\alpha \nabla_{\boldsymbol\theta}~J({\boldsymbol\theta}_k)$ and $\boldsymbol\phi_{k+1}~\leftarrow~\boldsymbol\phi_k~+~\alpha~\nabla_{\boldsymbol\phi}~L(\boldsymbol\phi_k)$ are identical in both the \ac{serl} and \ac{sprl} framework at every iteration $k\geq0$.
\end{lemma}
The proof is provided in appendix \ref{app:proof_lemma_1}. 
\begin{corollary}
    Under the conditions of Lemma \ref{lemma:serl_sprl_stochastic}, the sequences of parameters ${({\boldsymbol\theta}_k, \boldsymbol\phi_k)}_{k=0}^{\infty}$ generated by \ac{serl} and \ac{sprl} are identical, and consequently both frameworks converge to the same policy and value function parameters.
\end{corollary}
In contrast, for algorithms using deterministic policies, \ac{serl} and \ac{sprl} have different learning equations and may thus converge to different locally optimal policies as shown next. 

\subsection{Differences for Deterministic Policies}\label{sec:action_aliasing}
\begin{figure*}[t]
    \centering
    \begin{subfigure}[t]{0.32\textwidth}
        \includesvg[width=\textwidth]{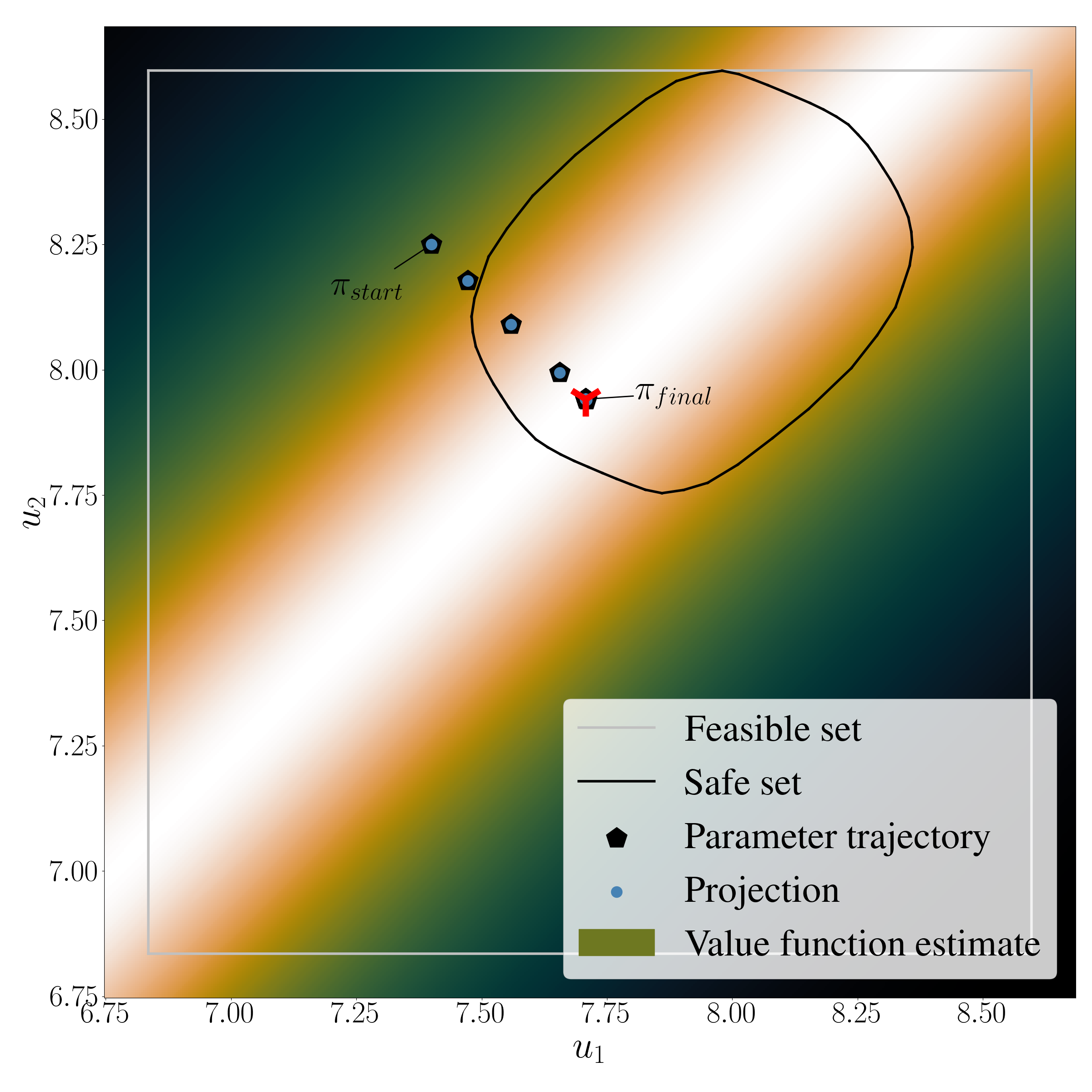}
        \caption{Without safeguarding, the critic approximates the true objective function, and the policy converges to the optimal safe action.}
        \label{fig:min_ex_unsafe}
    \end{subfigure}
    \begin{subfigure}[t]{0.32\textwidth}
        \includesvg[width=\textwidth]{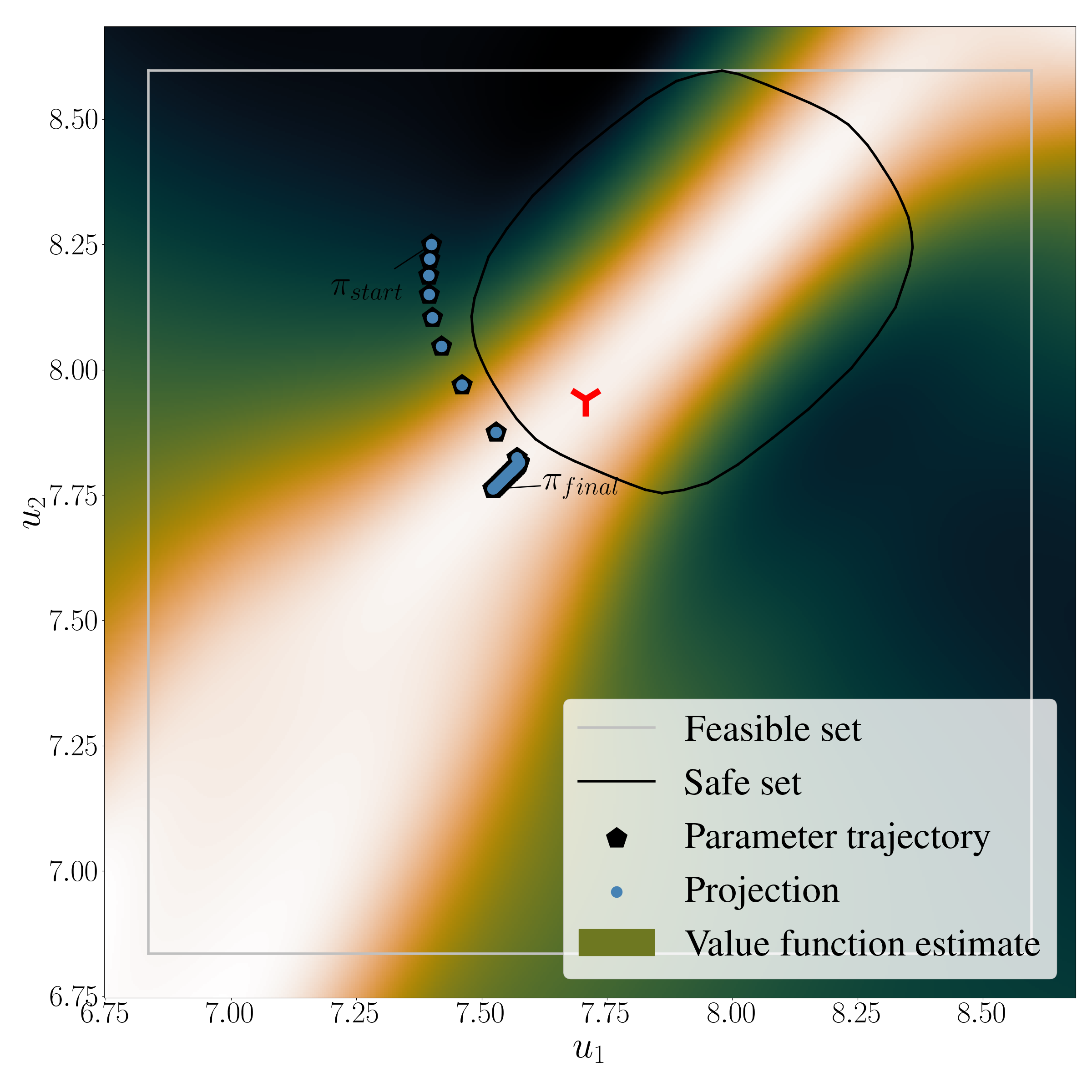}
        \caption{\ac{serl}: Policy converges to an unsafe and suboptimal action that lies in the direction normal to the safe optimal action and the safe set boundary.}
        \label{fig:min_ex_serl}
    \end{subfigure}    
    \begin{subfigure}[t]{0.32\textwidth}
        \includesvg[width=\textwidth]{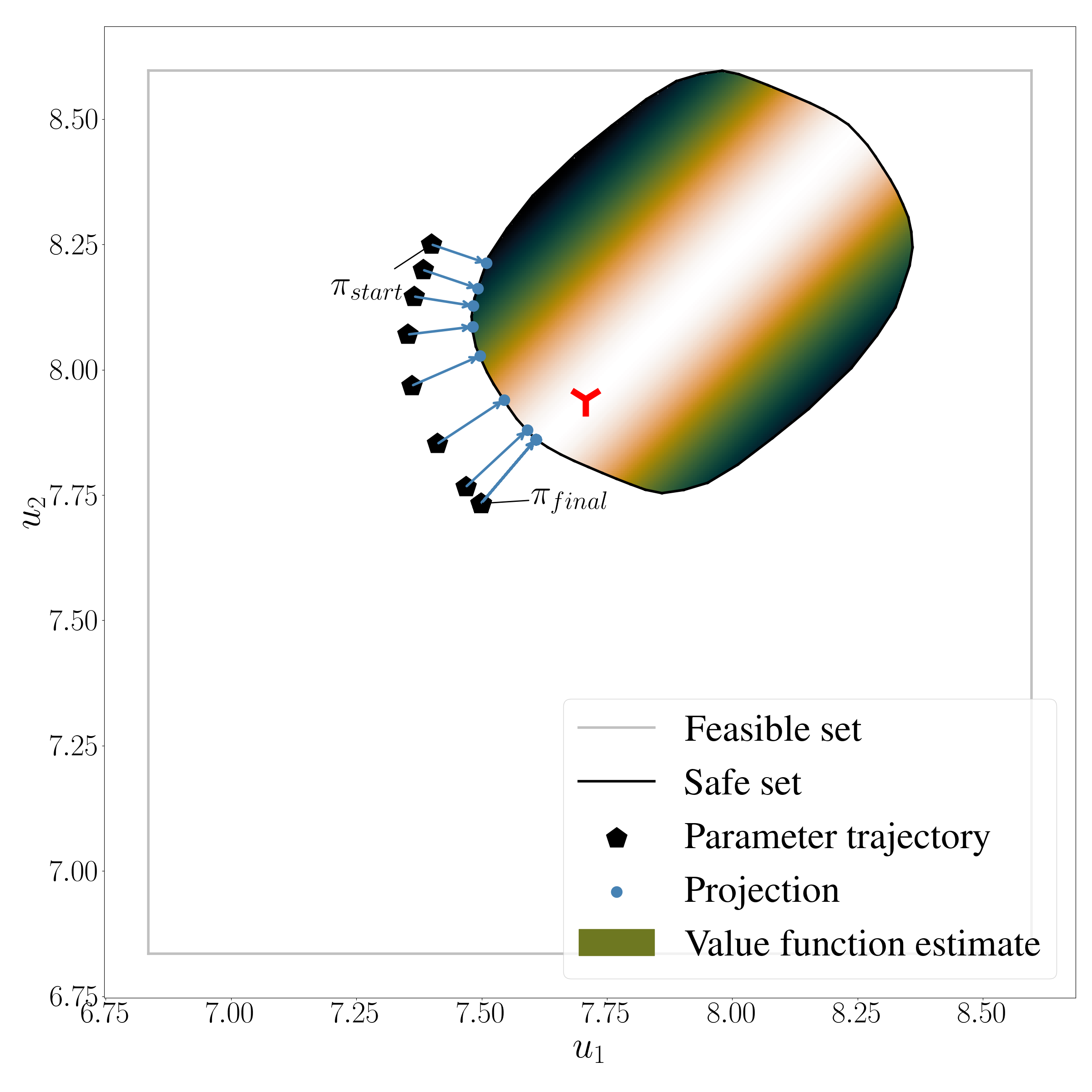}
        \caption{\ac{sprl}: Policy with differentiable projection safeguard does not improve in the direction normal to the projection and therefore does not reach the optimal safe action.}
        \label{fig:min_ex_sprl}
    \end{subfigure}
    \caption{Effect of action aliasing on \ac{serl} and \ac{sprl} algorithms using deterministic policies.  We illustrate the policy improvement step for a given state $\vx$ in the quadrotor balancing task. The deterministic policy $\pi(\vx)$ is updated for 50 steps using a loss function based on the learned state-action value function $q(\vx, \vu)$ in the \ac{serl} case and $q(\vx, \vu^\varphi)$ in the \ac{sprl} case.} 
    \label{fig:min_example}
\end{figure*}

As summarized in table \ref{tab:overview_equations}, the first difference consists in the scope of the critic.  Most algorithms using deterministic policies learn a state-action value function $q$. In \ac{sprl}, $q$ is evaluated only on safe actions and, thus, does not learn a meaningful approximation outside of the safe action set. In \ac{serl}, $q$ is also evaluated on unsafe actions. However, any value function in \ac{serl} that is conditioned on actions is affected by what we refer to as action aliasing: 
\begin{lemma}[Action aliasing]\label{def:action_aliasing}
   Let $\partial \sU^\varphi_\vx$ be the boundary of a convex safe action set $\sU^\varphi_\vx$ and $N_{\sU^\varphi_\vx}(\vu^b) = \{\vn \,|\, \vn^T(\vu^\varphi - \vu^b) \leq 0 \quad \forall \vu^\varphi \in \sU^\varphi_\vx\}$ the normal cone at $\vu^b \in \partial \sU^\varphi_\vx$ \citep{boyd2004convex}. Furthermore, let $\sU^e (\vu^b) = \{\vu^e \in \sU \setminus \sU^\varphi_\vx \,|\, \vu^e = \vu^b + \zeta \vn, \zeta > 0\, , \vn \in N_{\sU^\varphi_\vx}(\vu^b)\}$ be the set of unsafe actions $\vu^e$ in the normal cone of a given $\vu^b \in \partial \sU^\varphi_\vx$. Then, any action $\vu^e \in \sU^e(\vu^b)$ will be projected to $\vu^b$, causing a transition to the same next state $\vx^{\prime}$ and yielding the same reward $r^b \sim p_r(r \,|\, \vx, \vu^b$).
\end{lemma}

This means that the critic cannot distinguish between the actions $\vu^e \in \sU^e(\vu^b)$, potentially hindering learning. We formalize this \textit{flat-lining critic} phenomenon in the following lemma, which is proven in appendix \ref{app:proof_lemma_3}:
\begin{lemma}[Flat-lining critic]\label{lemma:flat_lining_critic}
    Following lemma \ref{def:action_aliasing}, the state-action value function, the advantage function, and the \ac{gae} adhere to 
    \begin{align*}
        q_\pi(\vx, \vu^e) &= q_\pi(\vx, \vu^b) \quad \forall \vu^e \in \sU^e, \\
        a_\pi(\vx, \vu^e) &= a_\pi(\vx, \vu^b) \quad \forall \vu^e \in \sU^e, \textnormal{ and} \\
        \hat{a}^{\text{GAE}}_\pi(\vx, \vu^e) &= \hat{a}^{\text{GAE}}_\pi(\vx, \vu^b) \quad \forall \vu^e \in \sU^e,
    \end{align*}
    respectively.
\end{lemma}
Note that the flat-lining critic issue also applies to algorithms using stochastic policies in \ac{serl}. Essentially, lemma \ref{lemma:flat_lining_critic} states that gradient information along the normal directions will be eliminated by using a perfect value function approximation in \eqref{eq:policy_gradient_estimate_determ}. 

We illustrate this using a simplified example. Consider the quadrotor task described in section \ref{app:benchmark_problems}. Here, we assume deterministic system dynamics and compute the safe action set $\sU^\varphi_\vx$ for a fixed state $\vx$ given the RCI set. We use supervised learning and a random behavior policy to train a critic network to approximate the state-action value function $q_\pi(\vx,\vu)$ for the given state. Then, we use a deterministic target policy represented by a tensor of the same dimension as $\vu$ and perform 50 policy update steps using stochastic gradient descent to maximize the learned objective. Note that for this environment, task performance (reaching the equilibrium state) and safety are well-aligned since the equilibrium is considered a safe state. Therefore, the optimal action is often also safe, as shown in figure \ref{fig:min_example}. In appendix \ref{app:min_example_seeker}, we provide a second example where task performance and safety are not aligned. 

Figure \ref{fig:min_ex_unsafe} shows a non-safeguarded setup, where the critic has learned a good approximation of the true objective function. The policy improvement steps lead to the optimal safe action. However, such a learning process is only possible in computer simulation or non-safety-critical environments, as unsafe actions might be executed. 
In \ac{serl}, the safeguarding takes place in the environment such that the projection is not visible in figure \ref{fig:min_ex_serl}. We observe that the policy actions never converge to the safe action set, and would not reach the optimal safe action even after the projection. This is caused by the flat-lining critic, which eliminates gradients in the direction normal to the boundary of the safe action set.

\ac{sprl} avoids the flat-lining critic by restricting evaluation to safe actions. We visualize this in figure \ref{fig:min_ex_sprl}. Instead, the impact of action aliasing is shifted to another part of the policy gradient estimation. When computing the sensitivity of the safeguard $\nabla_\vu\Phi(\vx,\vu)$, any components in the normal direction to the boundary vanish \citep{gros2020safe}, leading to a rank-deficient Jacobian. This is termed the \textit{zero-gradient problem} in existing work \citep{lin2021escaping, kasaura2023benchmarking}. 
As shown in figure \ref{fig:min_ex_seeker_sprl}, the zero-gradient problem can be particularly challenging for safe action sets with non-smooth surfaces, as multiple constraints are active on vertices, further reducing the rank of the Jacobian. This is also visible to a lesser extent for the flat-lining critic in figure \ref{fig:min_ex_seeker_serl}. 

We would like to highlight that while the zero-gradient problem in \ac{sprl} always exists, the flat-lining critic problem in \ac{serl} depends on the quality of the learned value function. An imperfect value function approximation may, in fact, alleviate the issue, as shown in other works highlighting the advantages of sampling-based gradient estimation \citep{suh2022differentiable}. We will examine the practical implications in section~\ref{sec:experiments}. 

\section{Addressing Action Aliasing in SE-RL and SP-RL}\label{sec:improvs}

We extend our comparative analysis to variants of \ac{serl} and \ac{sprl} that aim to address the shared issue of action aliasing. First, we establish how existing approaches differ, impairing the theoretical equivalence established in theorem \ref{theorem:eq_opt_val_functions}. Then, we propose an alternative approach for \ac{sprl} that re-establishes the validity.

\subsection{SE-RL}\label{sec:improv_serl}
\begin{figure*}[t]
    \centering
    \begin{subfigure}[t]{0.32\textwidth}
        \includesvg[width=\textwidth]{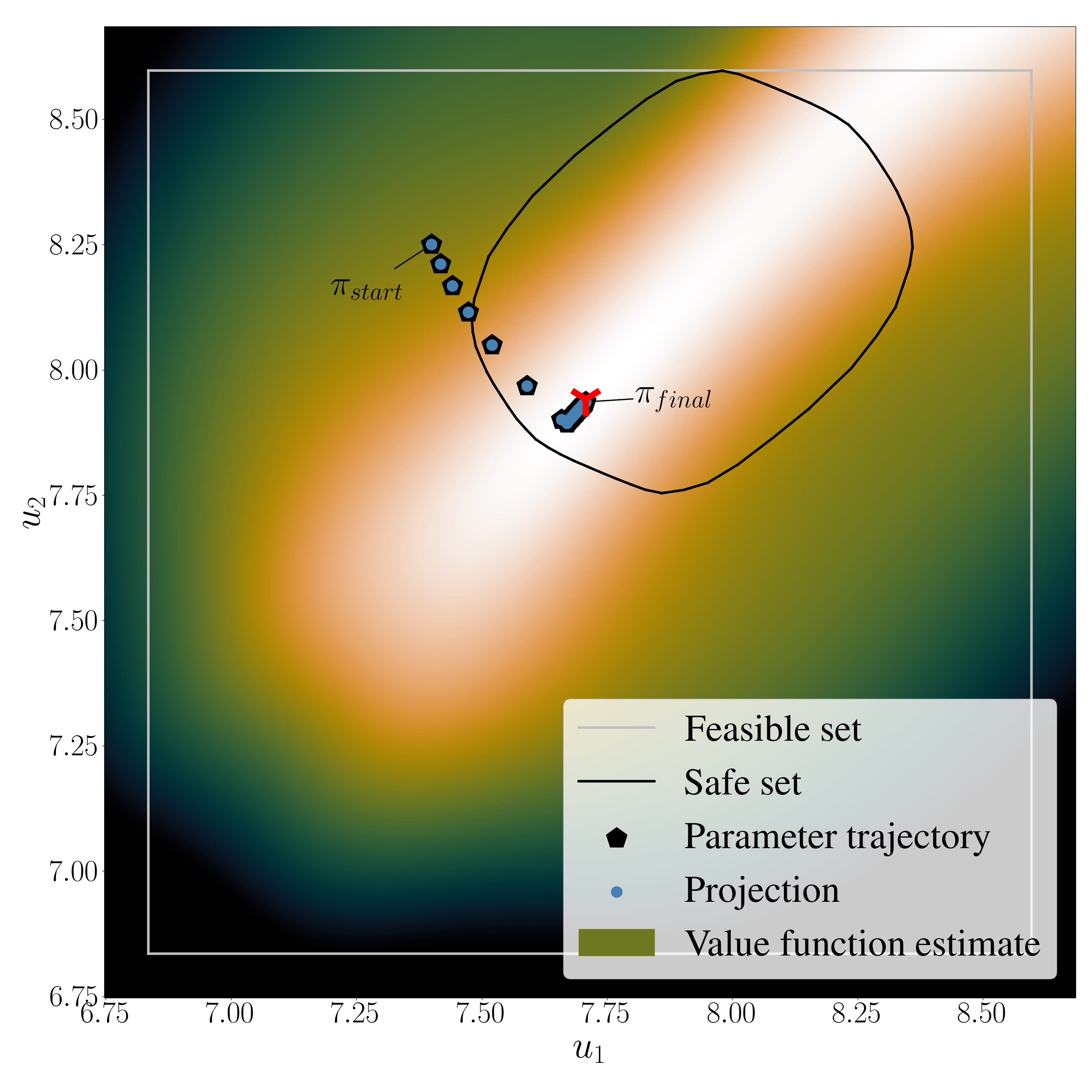}
        \caption{\ac{serl}: Adding a penalty to the reward addresses the flat-lining critic problem and improves convergence toward the optimal safe action.}
        \label{fig:min_ex_serl_pen}
    \end{subfigure}
    \begin{subfigure}[t]{0.32\textwidth}
        \includesvg[width=\textwidth]{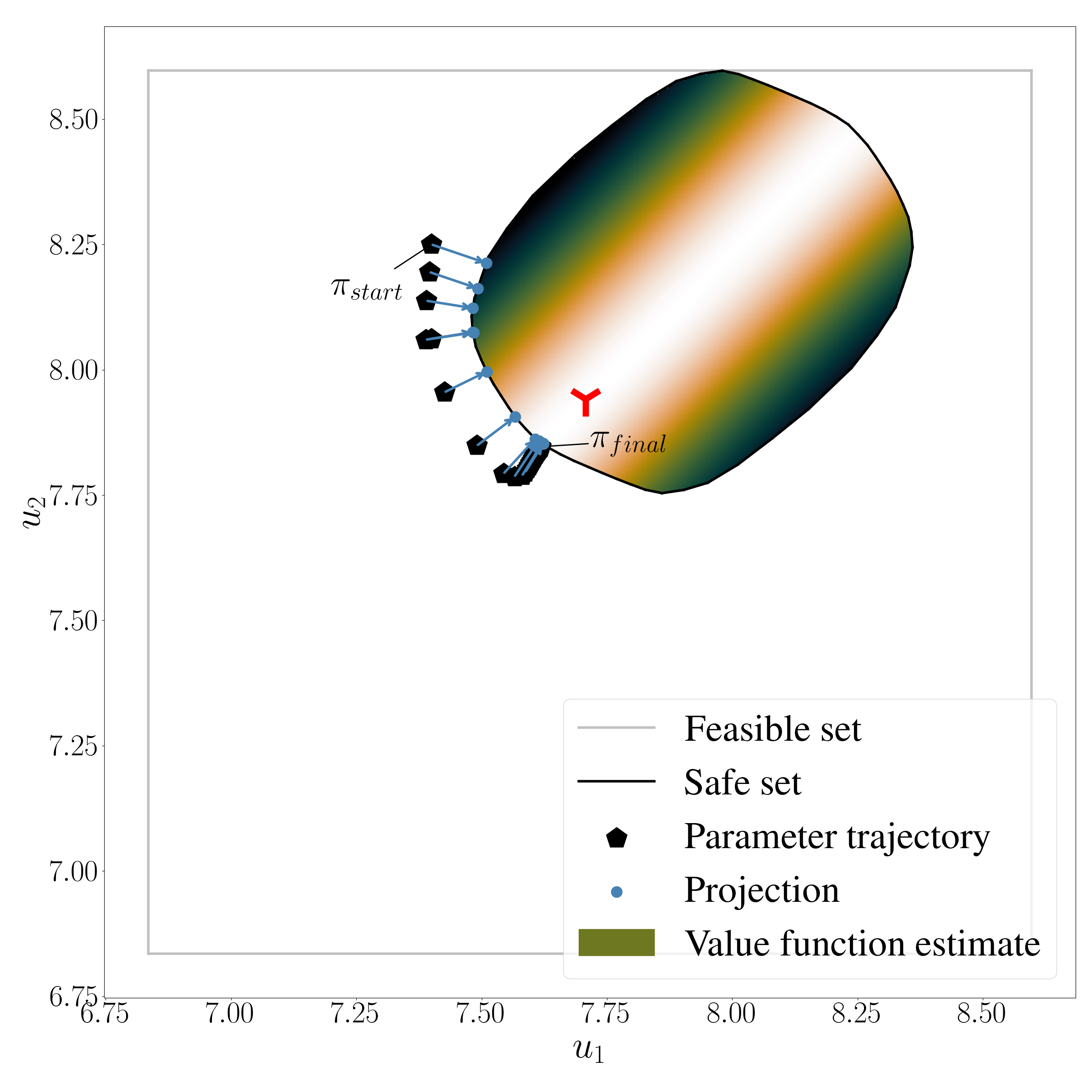}
        \caption{\ac{sprl}: An additional policy loss term that penalizes the distance between unsafe and safe action improves convergence to the optimal safe action but stops on the boundary of the safe action set.}
        \label{fig:min_ex_sprl_loss}
    \end{subfigure}    
    \begin{subfigure}[t]{0.32\textwidth}
        \includesvg[width=\textwidth]{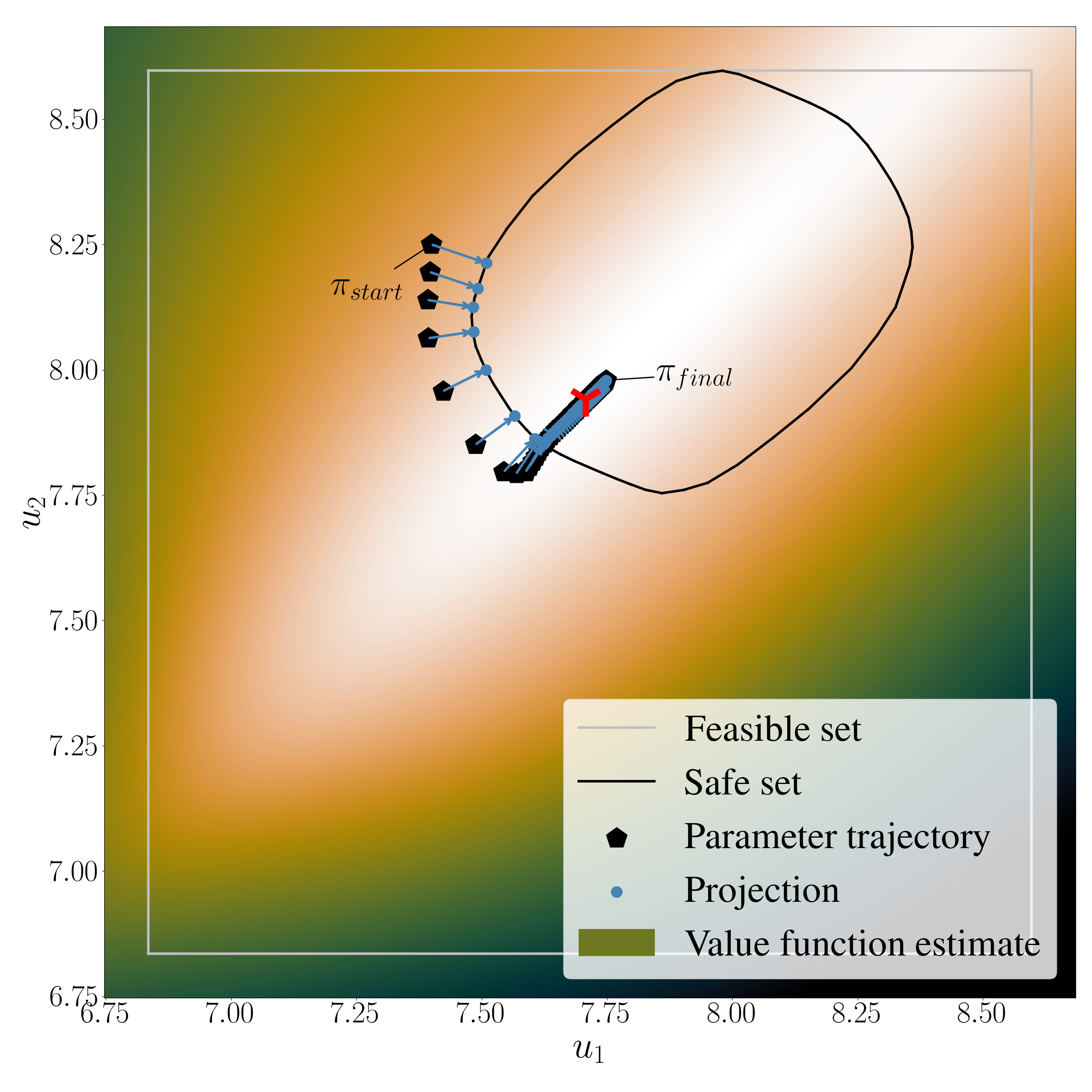}
        \caption{\ac{sprl}: Learning an additional penalty critic is very similar to adding a penalty to the reward in \ac{serl}. Note that as the penalty critic is conditioned on unsafe actions, we display the objective over the entire action space.}
        \label{fig:min_ex_sprl_critic}
    \end{subfigure}
    \caption{Effect of improvement strategies when using a differentiable safeguard during policy updates for a given state $\vx$ in the navigation task. The deterministic policy $\pi_{\boldsymbol\theta}(\vx)$ is updated for 50 steps using a loss function based on the learned state-action value function $q(\vx, \vu)$ in the \ac{serl} case and $q(\vx, \vu^\varphi)$ in the \ac{sprl} case.}. 
    \label{fig:min_example_improved}
\end{figure*}

In \ac{serl}, action aliasing is commonly addressed by adding a penalty $h$ to the reward each time the safeguard has to intervene \citep{wabersich2021predictive, markgraf2023safe, stanojev2023safe, bejarano2024safety, kasaura2023benchmarking, dawood2025safe}. This changes the \ac{mdp} reward function from section \ref{sec:serl} to $r^\text{SE,aug}: \tilde{\sX} \times \sU \times \sR \rightarrow \sR_{\geq 0}$ where
\begin{equation}
    r^\text{SE,aug}(\vx, \vu) = r^\text{SE}(\vx, \vu) - h.
\end{equation}
A common choice for the penalty function is the squared Euclidean distance between the safe and the unsafe action 
\begin{equation}\label{eq:squared_distance}
    h = \xi(\vu^\varphi, \vu) = \begin{cases}
        0 & \text{if } \vu \in \sU^\varphi_\vx, \\
        w \|\vu - \vu^\varphi\|_2^2 & \text{otherwise},
    \end{cases}
\end{equation}
where $w$ is a hyperparameter.
Introducing a penalty changes the optimization goal of \ac{serl} as shown in figure \ref{fig:min_ex_serl_pen}. To enable a better comparison with the improvement strategies for \ac{sprl} presented in section \ref{sec:improv_sprl}, we quantify the impact the penalty has on the value function and policy gradient estimate.  
The state-action value function $q_\pi^\text{aug}(\vx,\vu) $ can be rewritten as 
\begin{align*}
    q^\text{aug}_\pi(\vx,\vu) &= \E_\pi \left[ \sum_{k=0}  ^\infty \gamma^k \rr_{t+k+1}^\text{SE,aug} \,\bigg|\, \rvx_t=\vx, \rvu_t=\vu \right] \\
    &= \E_\pi \left[ \sum_{k=0}^\infty \gamma^k (\rr_{t+k+1}^\text{SE} - h_{t+k+1}) \,\bigg|\, \rvx_t=\vx, \rvu_t=\vu \right] \\
    &= \underbrace{\E_\pi \left[ \sum_{k=0}^\infty \gamma^k \rr_{t+k+1}^\text{SE} \,\bigg|\, \rvx_t=\vx, \rvu_t=\vu \right]}_{q^\text{SE}_\pi(\vx,\vu)} - \underbrace{\E_\pi \left[ \sum_{k=0}^\infty \gamma^k h_{t+k+1} \,\bigg|\, \rvx_t=\vx, \rvu_t=\vu  \right]}_{q^\text{pen}_\pi(\vx,\vu)}.
\end{align*}

Then, the policy gradient estimate for a deterministic policy can be expressed as
\begin{equation}\label{eq:aug_det_policy_grad}
    \nabla_{\boldsymbol\theta} J(\pi) = \E_\pi \left[ \nabla_{\boldsymbol\theta} \pi(\vx) \nabla_\vu q^\text{aug}_\pi(\vx,\vu)\right] = \E_\pi \left[\nabla_{\boldsymbol\theta} \pi(\vx) \nabla_\vu \left(q^\text{SE}_\pi(\vx, \vu) -  q^\text{pen}_\pi(\vx,\vu)\right)\right],
\end{equation}
showing the impact of the penalty. The term $q^\text{pen}_\pi(\vx,\vu)$ steers the policy away from actions that frequently trigger the safeguard, such that policy actions that are inherently safe or closer to the safe region are preferred. The strength of this effect is controlled by the hyperparameter $w$ in \eqref{eq:squared_distance}.
A similar derivation can be done for stochastic policies, resulting in 
\begin{equation}\label{eq:aug_stoch_policy_grad}
    \nabla_{\boldsymbol\theta} J(\pi) = \E_{\pi} \left[\Psi_\pi^\text{aug}(\vx,\vu) \nabla_{\boldsymbol\theta} \log \pi(\vu \,|\, \vx) \right] = \E_{\pi} \left[(\Psi_\pi^\text{SE}(\vx,\vu) - \Psi_\pi^\text{pen}(\vx,\vu)) \nabla_{\boldsymbol\theta} \log \pi(\vu \,|\, \vx) \right].
\end{equation}

\subsection{SP-RL}\label{sec:improv_sprl}
In \ac{sprl}, action aliasing manifests as rank-deficient Jacobians of the safeguard that directly enter the policy gradient computation through backpropagation. Unlike in \ac{serl}, where penalties can influence the value function approximation to mitigate action aliasing effects, reward penalties in \ac{sprl} cannot eliminate the underlying rank-deficiency in the sensitivity of the safeguard. Instead, existing literature in \ac{sprl} suggests a direct modification of the policy loss to improve performance \citep{chen2021enforcing, bhatia2019resource}. For deterministic policies, this results in a combined policy gradient estimate  
\begin{equation}\label{eq:det_grad_estimate_add_loss}
    \nabla_{\boldsymbol\theta} J(\pip) = \E_{\pip} \left[\nabla_{\boldsymbol\theta} \pip(\vx) \nabla_{\vu^\varphi} q^\text{SP}_{\pip}(\vx, \vu^\varphi) - \nabla_{\boldsymbol\theta} d(\cdot) \right],
\end{equation}
where $d(\cdot)$ commonly depends on the unsafe and the safe action. 
For example, in \citet{chen2021enforcing}, the authors suggest using the squared distance between the unsafe and the projected action as an additional loss such that $d=\xi(\pip(\vx), \pi(\vx))$ as defined in \eqref{eq:squared_distance}. Others use a loss term proportional to the constraint violation, which is theoretically similar \citep{bhatia2019resource}. We illustrate the effect of the squared distance loss in figure \ref{fig:min_ex_sprl_loss} for a deterministic policy. Compared to figure \ref{fig:min_ex_sprl}, the policy evolves toward the boundary of the safe action set, even though it does not reach the optimal safe action.
For stochastic policies, the vanishing gradient problem does not apply as discussed in section \ref{sec:action_aliasing}. Nevertheless, we can apply this additional loss to the mean $\vmu$ of the Gaussian policy $\mathcal
N(\vmu, \mSigma)$ and its projection $\vmu^\perp$ such that
\begin{equation}\label{eq:stoch_grad_estimate_add_loss}
    \nabla_{\boldsymbol\theta} J(\pip) = \E_{\pip} \left[\Psi_{\pip}(\vx,\vu^\varphi) \nabla_{\boldsymbol\theta} \log \pi(\vu \,|\, \vx) \right] - \E_{\pip}[\nabla_{\boldsymbol\theta} \xi(\vmu^\perp(\vx), \vmu(\vx))].
\end{equation}

Using an additional loss term as in \eqref{eq:det_grad_estimate_add_loss} and \eqref{eq:stoch_grad_estimate_add_loss} captures only myopic effects of unsafe actions. In contrast, penalties added to the reward are embedded into the value function estimate, providing information regarding their long-term consequences. Consequently, the per-sample loss in \ac{sprl} targets a different optimal value function than the improved \ac{serl} version presented in section \ref{sec:improv_serl}, breaking the equivalence established in theorem \ref{theorem:eq_opt_val_functions}. Furthermore, the per-sample loss can only push unsafe actions to the boundary of the safe action set, but not to its interior. 

Therefore, we suggest an alternative that consists of training an additional critic conditioned on the unsafe policy such that
\begin{equation}
    q^\text{pen}_\pi(\vx, \vu) = \E_\pi\left[\rg_t^\text{pen} \,|\, \rvx_t=\vx, \rvu_t=\vu\right],
\end{equation}
where $\rg_t^\text{pen} = \sum_{k=0}^\infty \gamma^k h_{t+k+1}$. This yields the deterministic policy gradient estimate
\begin{equation}\label{eq:sprl_improved_grad_det}
        \nabla_{\boldsymbol\theta} J(\pip) = \E_{\pip} \left[\nabla_{\boldsymbol\theta} \pip(x) \nabla_{\vu^\varphi} q^\text{SP}_{\pip}(\vx, \vu^\varphi)\right] - \E_{\pi} \left[\nabla_{\boldsymbol\theta} \pi(\vx) \nabla_\vu q^\text{pen}_\pi(\vx, \vu) \right],
\end{equation}
which is similar in structure to the one from \eqref{eq:aug_det_policy_grad}. The main difference to \ac{serl} is that the first term in \eqref{eq:sprl_improved_grad_det} depends on the safeguarded policy and involves the sensitivity of the safeguard. Since the second term is computed for the unsafe policy, convergence behavior compared to the vanilla loss function is improved as shown in figure \ref{fig:min_ex_sprl_critic}. In comparison to the per-sample loss, the policy reaches the interior of the safe action set.

In principle, this approach can also be applied to algorithms with stochastic policies, resulting in the policy gradient estimate
\begin{equation}
    \nabla_{\boldsymbol\theta} J(\pip) = \E_{\pip} \left[\Psi_{\pip}(\vx,\vu^\varphi) \nabla_{\boldsymbol\theta} \log \pi(\vu \,|\, \vx) \right] - \E_\pi \left[\Psi_\pi^\text{pen}(\vx,\vu) \nabla_{\boldsymbol\theta} \log \pi(\vu \,|\, \vx) \right]. 
\end{equation}

However, as discussed in section \ref{sec:sprl_stochastic_policies}, the distinction to \ac{serl} with additional penalties vanishes when the algorithm is based on learning a state value function $v^\text{SE}_{\pip}(\vx)$ that does not depend on the action. More details on the penalty critic implementation can be found in appendix \ref{app:penalty_critic_details}.

\section{Experiments}\label{sec:experiments}
We design our numerical experiments\footnote{Our implementation is available at: \href{https://github.com/TUMcps/serl-sprl}{https://github.com/TUMcps/serl-sprl}.} to answer the following questions: (1) Without any modifications, which approach achieves better returns empirically - \ac{serl} or \ac{sprl}? (2) How do the approaches for addressing action aliasing in \ac{serl} and \ac{sprl} compare to each other? For \ac{sprl}, we consider the per-sample squared distance loss (PSL) and our proposed penalty critic (PenC), and for \ac{serl} proportional penalties.

We restrict our investigation to actor-critic algorithms based on the policy gradient theorem, as \citet{gros2020safe} provide a derivation of unbiased policy gradient estimates under the \ac{sprl} framework for these algorithms. We choose \ac{td3} \citep{fujimoto2018addressing} and \ac{a2c} \citep{mnih2016asynchronous} as RL algorithms using deterministic and stochastic policies, respectively. 

\begin{figure}[t]
    \centering
    \begin{subfigure}[t]{\textwidth}
    \includesvg[width=\linewidth]{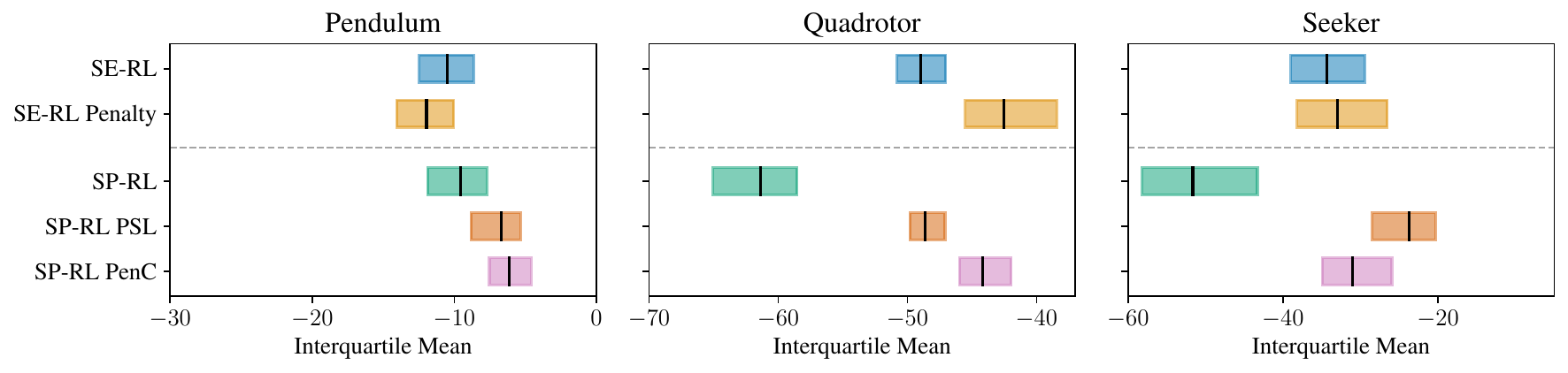}
    \caption{Results for \ac{td3} (deterministic policies).}
    \label{fig:iqm_td3}
    \end{subfigure} \\
    \begin{subfigure}[t]{\textwidth}
    \includesvg[width=\linewidth]{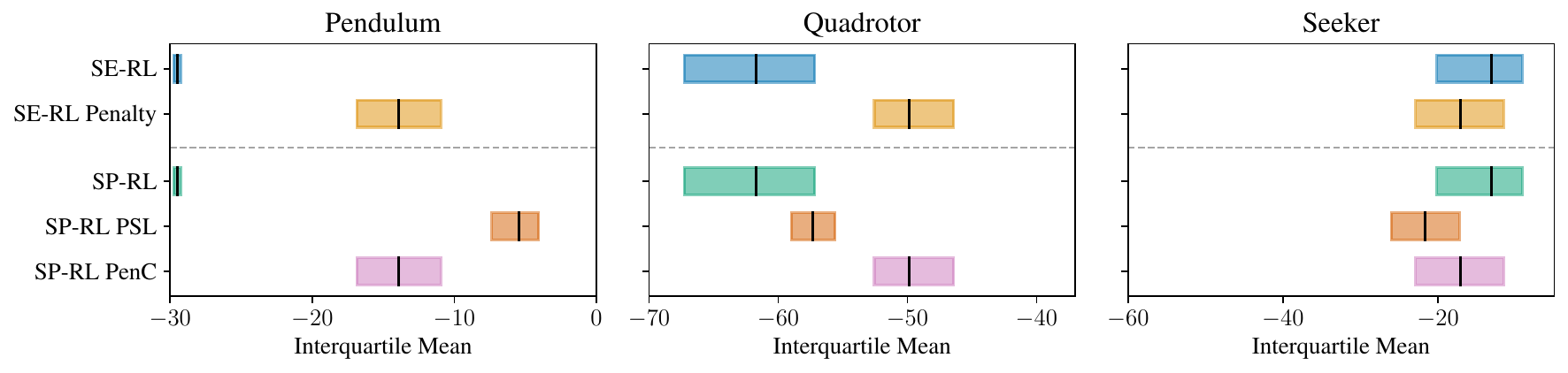}
    \caption{Results for \ac{a2c} (stochastic policies). For the pendulum, we clip the returns obtained with vanilla A2C to enable a better comparison. The actual returns are much lower as shown in table \ref{tab:pend_a2c}.}
    \label{fig:iqm_a2c}
    \end{subfigure}
    \caption{Comparison of vanilla and modified \ac{serl}/\ac{sprl} approaches (PSL: per-sample loss, PenC: penalty critic). We provide the interquartile mean and the 95\% confidence interval of the return achieved at test time. For the improved \ac{serl}/\ac{sprl} approaches, we only show the result for the best-performing choice of $w \in \{0.1, 0.5, 1.0, 2.0\}$. The full results are listed in appendix \ref{app:full_results}.}
    \label{fig:deployment_results}
\end{figure}

We use cvxpylayers \citep{agrawal2019differentiable} to integrate the safeguard into the policy for \ac{sprl} as it automates the computation of the sensitivities of the projection in the backward pass. To ensure a fair comparison, we run all experiments on the same CPU (Intel Core i9-14900K). Furthermore, we conduct hyperparameter tuning for the unsafe baselines of each algorithm and then use this set of hyperparameters for all experiments. The weighting factor $w$ is an important hyperparameter. Therefore, we do not tune it, but instead test for different choices: $w \in \{0.1, 0.5, 1.0, 2.0\}$. We report the final performance using the interquartile mean and 95\% bootstrapped confidence intervals of the undiscounted return over 7 training runs evaluated on 10 random seeds, respectively. 

\subsection{Benchmark Problems}\label{sec:benchmarks}
We perform the evaluation using a stabilization task for two classic control examples -- a pendulum and a quadrotor -- as well as a navigation task, and an energy management system (EMS) optimization task. For the first two tasks, we compute RCI sets using the approach from \citet{schafer2023scalable}. Since the RCI sets are centered around the equilibrium point, task performance is closely aligned with safety. This is different in the navigation task, where a simple point mass seeker has to find the shortest path to a goal while avoiding obstacles, such that the optimal safe policy operates in close proximity to the unsafe regions. Here, instead of an RCI set, we use a state-dependent safe action set to avoid collisions. In the EMS task, there are two competing objectives: maintaining the indoor temperature close to a set point and minimizing electricity cost. Consequently, the alignment of safety and performance depends on which objective dominates at a given point in time. Further details on the benchmark problems are provided in appendix \ref{app:benchmark_problems}.

\subsection{Results}

\subsubsection{Comparison of SE-RL and SP-RL Without Modifications}
As shown in lemma \ref{lemma:serl_sprl_stochastic}, for algorithms such as \ac{a2c} that employ stochastic policies and learn a state value function, \ac{serl} and \ac{sprl} result in the same policies. Consequently, they deliver the same return at test time as shown in figure \ref{fig:iqm_a2c}. 

For \ac{td3}, figure \ref{fig:iqm_td3} shows that vanilla \ac{serl} often outperforms \ac{sprl}, especially for more complex environments. For the pendulum task, the final performance is very similar, with a slight advantage for \ac{sprl}. For the quadrotor, the EMS, and the seeker task, the differences are more pronounced, and vanilla \ac{serl} clearly dominates \ac{sprl}. Figure \ref{fig:return_td3} shows a large variance in the training performance for \ac{td3}-\ac{sprl} caused by several non-convergent runs, which explains the poor performance at test time. 

\subsubsection{Comparison of Improved SE-RL and SP-RL}
\begin{figure}[t]
    \centering
    \includesvg[width=\linewidth]{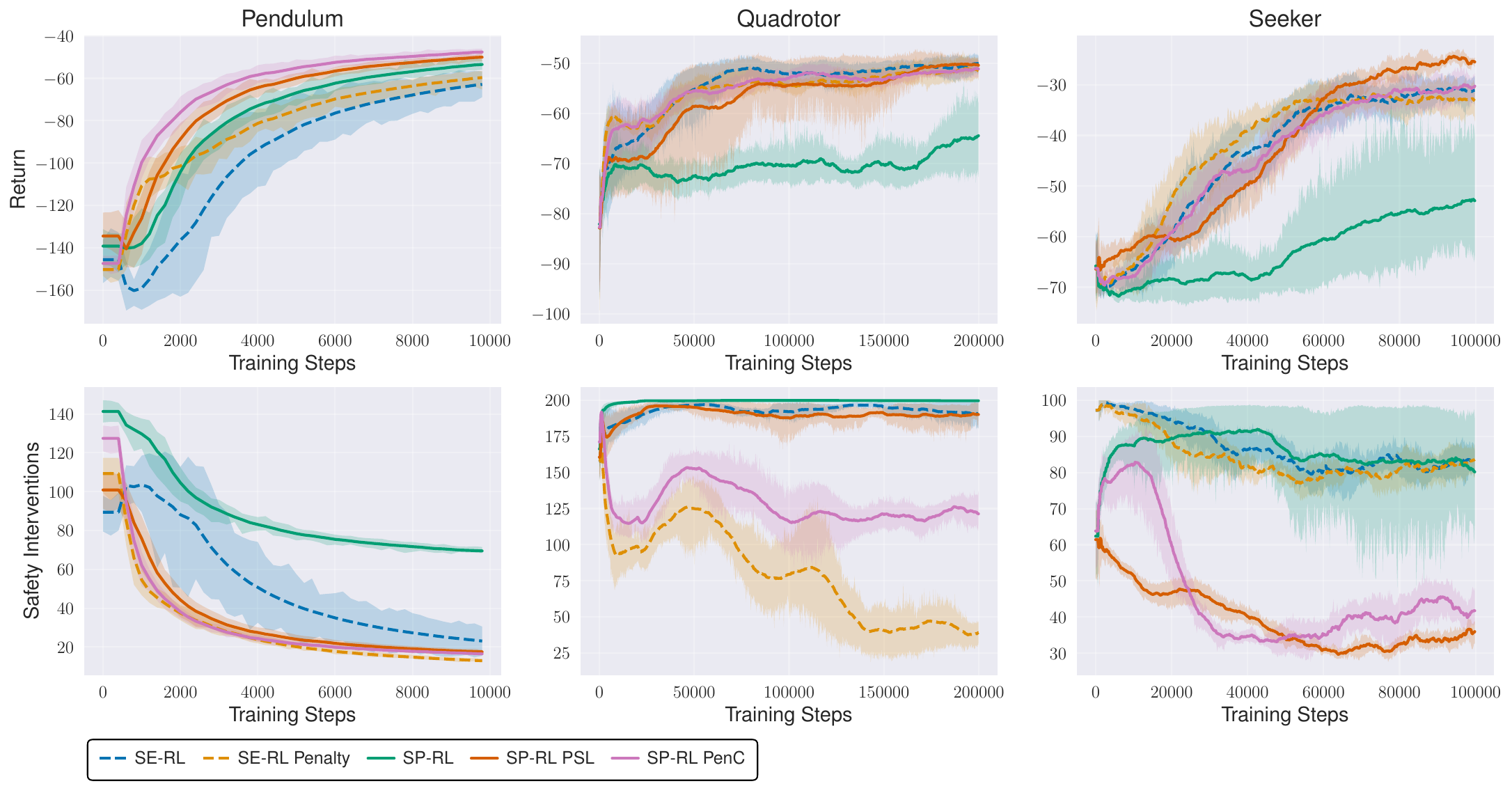}
    \caption{Interquartile mean and 95\% bootstrap confidence interval of the return and safeguard interventions over environment steps during training with \ac{td3}. We compare the vanilla \ac{serl} and \ac{sprl} versions and the strategies for mitigating action aliasing.}
    \label{fig:return_td3}
\end{figure}

In figure \ref{fig:deployment_results}, we only report the results for the best-performing choice of $w$, while the full results are shown in appendix \ref{app:full_results}. For the quadrotor and the seeker environment, the choices of $w$ that perform well are largely consistent for one algorithm. For the pendulum, larger $w$ in \ac{serl} can deteriorate performance significantly, while \ac{sprl} with the per-sample loss is very robust. 

For \ac{td3}-\ac{sprl}, both the per-sample loss and the penalty critic improve test performance across all four environments (see figure \ref{fig:iqm_td3}). As shown in figure \ref{fig:return_td3}, the convergence issues of vanilla \ac{td3}-\ac{sprl} in the quadrotor, the EMS, and the seeker can be mitigated. Which of the two strategies for action aliasing performs better depends on the type of environment: In the two balancing tasks and the EMS, the penalty critic delivers better results, while the per-sample loss has the edge in the navigation task. Statistical analysis using Kruskal-Wallis and Dunn's tests with Bonferroni correction ($p < 0.05$) confirmed that compared to \ac{td3}-\ac{serl} with penalties, one of the improved \ac{sprl} versions always performs on par or better.

For \ac{a2c}, the penalty in \ac{serl} and the penalty critic in \ac{sprl} are again equivalent. They significantly improve performance for the pendulum and quadrotor environments, and slightly for the EMS. In the seeker environment, performance slightly deteriorates for all improvement strategies in both \ac{serl} and \ac{sprl}. 

\section{Discussion}\label{sec:discussion}
For actor-critic algorithms using stochastic policies and \ac{gae}, both theoretical and empirical results confirm that \ac{serl} and \ac{sprl} are equivalent. While the per-sample loss on the policy mean can sometimes improve performance, it is usually outperformed by penalty-based approaches. The seeker environment proved more challenging for improvement strategies due to the difficulty of weighing task performance with safety and the already strong performance of vanilla \ac{a2c}.

As reported previously \citep{pham2018optlayer}, we observe convergence issues when using \ac{sprl} with deterministic policies, particularly in complex environments, while \ac{serl} shows no such issues. This supports our theoretical finding that critic flat-lining only hinders learning under perfect value function approximations. In contrast, rank-deficient Jacobians always affect policy updates in \ac{sprl}, making action aliasing more detrimental. Following \citet{bhatia2019resource, chen2021enforcing}, we find that these convergence issues can be mitigated by adding loss terms proportional to the distance between safe and unsafe actions.

The relative effectiveness of our proposed penalty critic versus the per-sample loss depends on the alignment between safety and task performance. In the balancing tasks and the EMS, where these objectives align, the penalty critic outperforms the per-sample loss because it can converge to optimal actions within the safe set, while the per-sample loss is limited to the boundary (figures \ref{fig:min_ex_sprl_loss} and \ref{fig:min_ex_sprl_critic}). In the seeker environment, where optimal task actions typically lie outside the safe set, both methods converge to boundary actions, making the per-sample loss more efficient.

Overall, improved \ac{sprl} variants typically match or exceed \ac{serl} performance for deterministic policies, but require 
$3$–$12$ times the computation time of \ac{serl} due to sensitivity computations via cvxpylayers. Our wall clock time analysis in appendix \ref{app:wall_clock_times} shows that the increase in computation time scales with the complexity of the projection problem (\ref{eq:projection_problem}). We observe the smallest difference between \ac{serl} and \ac{sprl} for the seeker task, where the constraints (\ref{eq:projection_constraints}) reflect a point containment problem given an unsafe action and a safe action set. This is much simpler than the set containment problem that has to be solved for the pendulum, the EMS, and the quadrotor task (see appendix \ref{app:zonotopes}). Since the state and action space dimensionality of the quadrotor and the EMS are higher than that of the pendulum, they feature the highest gap between \ac{serl} and \ac{sprl} wall clock times. While for such environments the performance gains in \ac{sprl} have to be weighed against computational complexity, we expect recent advances in the field of differentiable optimization layers \citep{nguyen2025fsnet, grontas2025pinet, frey2025differentiable} to mitigate this issue in the future. Furthermore, it is often possible to decouple high-dimensional safety-critical dynamics into several subproblems, which can then be solved in parallel during both the forward and the backward pass \citep{chen2021enforcing}.

\section{Conclusion}\label{sec:conclusion}
We present a comprehensive theoretical and empirical comparison of \ac{serl} and \ac{sprl}, two prominent approaches for integrating projection-based safeguards into actor-critic \ac{rl}. Our unified formalization enables us to prove that both approaches share optimal value functions but differ in their learning dynamics. For a specific subclass of \ac{rl} algorithms that use stochastic policies and \ac{gae}, we show the theoretical and practical equivalence of \ac{serl} and \ac{sprl}. For algorithms using deterministic policies, the approaches diverge depending on whether the sensitivity of the safeguard mapping is used explicitly in the backward pass of the policy loss (\ac{sprl}), or whether it is learned implicitly through the value function approximation (\ac{serl}). A central concept when analyzing these differences is action aliasing, where multiple unsafe actions are mapped to identical safe actions. Our analysis reveals that the effect of action aliasing is more detrimental to \ac{sprl} than \ac{serl}, potentially leading to convergence issues. We propose a novel penalty critic for \ac{sprl} that estimates discounted cumulative penalties proportional to the distance between safe and unsafe actions, providing a principled mitigation strategy aligned with penalty-based approaches in \ac{serl}. Our theoretical findings and empirical analysis provide practitioners with clear guidance: Vanilla \ac{serl} presents a strong baseline, particularly for environments where scaling additional penalties or loss terms is challenging, while improved \ac{sprl} variants should be considered when performance gains justify the additional computational cost.
\newpage

\subsubsection*{Acknowledgments}
The authors gratefully acknowledge the partial financial support of this work by the German Research Foundation through the SAFARI project (grant no. 458030766) and the SFB 1608 (grant no. 501798263), the Research Council of Norway (grant no. 300172, SARLEM), and the research training group ConVeY, funded by the German Research Foundation under grant GRK 2428/2. 
Furthermore, we thank Jonathan K{\"u}lz for his ideas on how to visualize the action aliasing phenomenon for the different learning paradigms. 

\bibliography{safe_learning}

\begin{thebibliography}{43}
\providecommand{\natexlab}[1]{#1}
\providecommand{\url}[1]{\texttt{#1}}
\expandafter\ifx\csname urlstyle\endcsname\relax
  \providecommand{\doi}[1]{doi: #1}\else
  \providecommand{\doi}{doi: \begingroup \urlstyle{rm}\Url}\fi

\bibitem[Agrawal et~al.(2019)Agrawal, Amos, Barratt, Boyd, Diamond, and Kolter]{agrawal2019differentiable}
Akshay Agrawal, Brandon Amos, Shane Barratt, Stephen Boyd, Steven Diamond, and J~Zico Kolter.
\newblock Differentiable convex optimization layers.
\newblock \emph{Advances in Neural Information Processing Systems}, 2019.

\bibitem[Bejarano et~al.(2025)Bejarano, Brunke, and Schoellig]{bejarano2024safety}
Federico~Pizarro Bejarano, Lukas Brunke, and Angela~P Schoellig.
\newblock Safety filtering while training: Improving the performance and sample efficiency of reinforcement learning agents.
\newblock \emph{IEEE Robotics and Automation Letters}, 10\penalty0 (1):\penalty0 788--795, 2025.

\bibitem[Bertsekas(2018)]{bertsekas2018dynamic}
Dimitri~P. Bertsekas.
\newblock \emph{Dynamic programming and optimal control}.
\newblock Athena Scientific, 2018.

\bibitem[Bhatia et~al.(2019)Bhatia, Varakantham, and Kumar]{bhatia2019resource}
Abhinav Bhatia, Pradeep Varakantham, and Akshat Kumar.
\newblock Resource constrained deep reinforcement learning.
\newblock In \emph{Proceedings of the International Conference on Automated Planning and Scheduling}, volume~29, pp.\  610--620, 2019.

\bibitem[Bianchi(2006)]{bianchi2006adaptive}
Mikael~Andreas Bianchi.
\newblock \emph{Adaptive modellbasierte pr{\"a}diktive Regelung einer Kleinw{\"a}rmepumpenanlage}.
\newblock PhD thesis, ETH Zurich, 2006.

\bibitem[Boyd \& Vandenberghe(2004)Boyd and Vandenberghe]{boyd2004convex}
Stephen Boyd and Lieven Vandenberghe.
\newblock \emph{Convex optimization}.
\newblock Cambridge University Press, 2004.

\bibitem[Chen et~al.(2021)Chen, Donti, Baker, Kolter, and Berg{\'e}s]{chen2021enforcing}
Bingqing Chen, Priya~L Donti, Kyri Baker, J~Zico Kolter, and Mario Berg{\'e}s.
\newblock Enforcing policy feasibility constraints through differentiable projection for energy optimization.
\newblock In \emph{Proceedings of the ACM International Conference on Future Energy Systems}, pp.\  199--210, 2021.

\bibitem[Dalal et~al.(2018)Dalal, Dvijotham, Vecerik, Hester, Paduraru, and Tassa]{dalal2018safe}
Gal Dalal, Krishnamurthy Dvijotham, Matej Vecerik, Todd Hester, Cosmin Paduraru, and Yuval Tassa.
\newblock Safe exploration in continuous action spaces.
\newblock \emph{arXiv preprint arXiv:1801.08757}, 2018.

\bibitem[Dawood et~al.(2025)Dawood, Pan, Dengler, Zhou, Schoellig, and Bennewitz]{dawood2025safe}
Murad Dawood, Sicong Pan, Nils Dengler, Siqi Zhou, Angela~P Schoellig, and Maren Bennewitz.
\newblock Safe multi-agent reinforcement learning for behavior-based cooperative navigation.
\newblock \emph{IEEE Robotics and Automation Letters}, 10\penalty0 (6):\penalty0 6256--6263, 2025.

\bibitem[Frey et~al.(2025)Frey, Baumgärtner, Frison, Reinhardt, Hoffmann, Fichtner, Gros, and Diehl]{frey2025differentiable}
Jonathan Frey, Katrin Baumgärtner, Gianluca Frison, Dirk Reinhardt, Jasper Hoffmann, Leonard Fichtner, Sebastien Gros, and Moritz Diehl.
\newblock Differentiable nonlinear model predictive control.
\newblock \emph{arXiv preprint arXiv:2505.01353}, 2025.

\bibitem[Fujimoto et~al.(2018)Fujimoto, Hoof, and Meger]{fujimoto2018addressing}
Scott Fujimoto, Herke Hoof, and David Meger.
\newblock Addressing function approximation error in actor-critic methods.
\newblock In \emph{International Conference on Machine Learning}, pp.\  1587--1596. Proceedings of Machine Learning Research, 2018.

\bibitem[Garc\'{\i}a \& Fern\'{a}ndez(2015)Garc\'{\i}a and Fern\'{a}ndez]{Garcia2015}
Javier Garc\'{\i}a and Fernando Fern\'{a}ndez.
\newblock A comprehensive survey on safe reinforcement learning.
\newblock In \emph{Journal of Machine Learning Research}, volume~16, pp.\  1437–1480, 2015.

\bibitem[Ghasemi et~al.(2020)Ghasemi, Sadraddini, and Belta]{ghasemi2020compositional}
Kasra Ghasemi, Sadra Sadraddini, and Calin Belta.
\newblock Compositional synthesis via a convex parameterization of assume-guarantee contracts.
\newblock In \emph{Proceedings of the International Conference on Hybrid Systems: Computation and Control}, pp.\  1--10, 2020.

\bibitem[Grontas et~al.(2026)Grontas, Terpin, Balta, D'Andrea, and Lygeros]{grontas2025pinet}
Panagiotis~D Grontas, Antonio Terpin, Efe~C Balta, Raffaello D'Andrea, and John Lygeros.
\newblock Pinet: Optimizing hard-constrained neural networks with orthogonal projection layers.
\newblock In \emph{International Conference on Learning Representations}, 2026.

\bibitem[Gros et~al.(2020)Gros, Zanon, and Bemporad]{gros2020safe}
Sebastien Gros, Mario Zanon, and Alberto Bemporad.
\newblock Safe reinforcement learning via projection on a safe set: How to achieve optimality?
\newblock \emph{IFAC-PapersOnLine}, 53\penalty0 (2):\penalty0 8076--8081, 2020.

\bibitem[Hunt et~al.(2021)Hunt, Fulton, Magliacane, Hoang, Das, and Solar-Lezama]{hunt2021verifiably}
Nathan Hunt, Nathan Fulton, Sara Magliacane, Trong~Nghia Hoang, Subhro Das, and Armando Solar-Lezama.
\newblock Verifiably safe exploration for end-to-end reinforcement learning.
\newblock In \emph{Proceedings of the International Conference on Hybrid Systems: Computation and Control}, pp.\  1--11, 2021.

\bibitem[Kasaura et~al.(2023)Kasaura, Miura, Kozuno, Yonetani, Hoshino, and Hosoe]{kasaura2023benchmarking}
Kazumi Kasaura, Shuwa Miura, Tadashi Kozuno, Ryo Yonetani, Kenta Hoshino, and Yohei Hosoe.
\newblock Benchmarking actor-critic deep reinforcement learning algorithms for robotics control with action constraints.
\newblock \emph{IEEE Robotics and Automation Letters}, 8\penalty0 (8):\penalty0 4449--4456, 2023.

\bibitem[Krantz \& Parks(2002)Krantz and Parks]{krantz2002implicit}
Steven~George Krantz and Harold~R Parks.
\newblock \emph{The implicit function theorem: history, theory, and applications}.
\newblock Springer Science \& Business Media, 2002.

\bibitem[Krasowski et~al.(2023)Krasowski, Thumm, M{\"u}ller, Sch{\"a}fer, Wang, and Althoff]{krasowski2023provably}
Hanna Krasowski, Jakob Thumm, Marlon M{\"u}ller, Lukas Sch{\"a}fer, Xiao Wang, and Matthias Althoff.
\newblock Provably safe reinforcement learning: Conceptual analysis, survey, and benchmarking.
\newblock \emph{Transactions on Machine Learning Research}, 2023.

\bibitem[Kulmburg \& Althoff(2021)Kulmburg and Althoff]{kulmburg2021co}
Adrian Kulmburg and Matthias Althoff.
\newblock On the co-{NP}-completeness of the zonotope containment problem.
\newblock \emph{European Journal of Control}, 62:\penalty0 84--91, 2021.

\bibitem[Lillicrap et~al.(2015)Lillicrap, Hunt, Pritzel, Heess, Erez, Tassa, Silver, and Wierstra]{lillicrap2015continuous}
Timothy~P Lillicrap, Jonathan~J Hunt, Alexander Pritzel, Nicolas Heess, Tom Erez, Yuval Tassa, David Silver, and Daan Wierstra.
\newblock Continuous control with deep reinforcement learning.
\newblock \emph{arXiv preprint arXiv:1509.02971}, 2015.

\bibitem[Lin et~al.(2021)Lin, Hung, Yang, Hsieh, and Liu]{lin2021escaping}
Jyun-Li Lin, Wei Hung, Shang-Hsuan Yang, Ping-Chun Hsieh, and Xi~Liu.
\newblock Escaping from zero gradient: Revisiting action-constrained reinforcement learning via {Frank-Wolfe} policy optimization.
\newblock In \emph{Uncertainty in Artificial Intelligence}, pp.\  397--407. Proceedings of Machine Learning Research, 2021.

\bibitem[Markgraf \& Althoff(2023)Markgraf and Althoff]{markgraf2023safe}
Hannah Markgraf and Matthias Althoff.
\newblock Safe multi-agent reinforcement learning for price-based demand response.
\newblock In \emph{IEEE PES Innovative Smart Grid Technologies Europe}, pp.\  1--6, 2023.

\bibitem[Marvi \& Kiumarsi(2022)Marvi and Kiumarsi]{marvi2022reinforcement}
Zahra Marvi and Bahare Kiumarsi.
\newblock Reinforcement learning with safety and stability guarantees during exploration for linear systems.
\newblock \emph{IEEE Open Journal of Control Systems}, 1:\penalty0 322--334, 2022.

\bibitem[Mnih et~al.(2016)Mnih, Badia, Mirza, Graves, Lillicrap, Harley, Silver, and Kavukcuoglu]{mnih2016asynchronous}
Volodymyr Mnih, Adria~Puigdomenech Badia, Mehdi Mirza, Alex Graves, Timothy Lillicrap, Tim Harley, David Silver, and Koray Kavukcuoglu.
\newblock Asynchronous methods for deep reinforcement learning.
\newblock In \emph{International Conference on Machine Learning}, pp.\  1928--1937, 2016.

\bibitem[Nguyen \& Donti()Nguyen and Donti]{nguyen2025fsnet}
Hoang~T Nguyen and Priya~L Donti.
\newblock {FSNet}: Feasibility-seeking neural network for constrained optimization with guarantees.
\newblock \emph{Advances in Neural Information Processing Systems}.

\bibitem[Pham et~al.(2018)Pham, De~Magistris, and Tachibana]{pham2018optlayer}
Tu-Hoa Pham, Giovanni De~Magistris, and Ryuki Tachibana.
\newblock Optlayer -- practical constrained optimization for deep reinforcement learning in the real world.
\newblock In \emph{IEEE International Conference on Robotics and Automation}, pp.\  6236--6243, 2018.

\bibitem[Sch{\"a}fer et~al.(2024)Sch{\"a}fer, Gruber, and Althoff]{schafer2023scalable}
Lukas Sch{\"a}fer, Felix Gruber, and Matthias Althoff.
\newblock Scalable computation of robust control invariant sets of nonlinear systems.
\newblock \emph{IEEE Transactions on Automatic Control}, 69\penalty0 (2):\penalty0 755--770, 2024.

\bibitem[Schervish(1995)]{schervish1995theory}
Mark~J Schervish.
\newblock \emph{Theory of statistics}.
\newblock Springer Science \& Business Media, 1995.

\bibitem[Schulman et~al.(2015)Schulman, Moritz, Levine, Jordan, and Abbeel]{schulman2015high}
John Schulman, Philipp Moritz, Sergey Levine, Michael Jordan, and Pieter Abbeel.
\newblock High-dimensional continuous control using generalized advantage estimation.
\newblock \emph{arXiv preprint arXiv:1506.02438}, 2015.

\bibitem[Schulman et~al.(2017)Schulman, Wolski, Dhariwal, Radford, and Klimov]{schulman2017proximal}
John Schulman, Filip Wolski, Prafulla Dhariwal, Alec Radford, and Oleg Klimov.
\newblock Proximal policy optimization algorithms.
\newblock \emph{arXiv preprint arXiv:1707.06347}, 2017.

\bibitem[Selim et~al.(2022)Selim, Alanwar, El-Kharashi, Abbas, and Johansson]{selim2022safe}
Mahmoud Selim, Amr Alanwar, M~Watheq El-Kharashi, Hazem~M Abbas, and Karl~H Johansson.
\newblock Safe reinforcement learning using data-driven predictive control.
\newblock In \emph{International Conference on Communications, Signal Processing, and their Applications}, pp.\  1--6, 2022.

\bibitem[Silver et~al.(2014)Silver, Lever, Heess, Degris, Wierstra, and Riedmiller]{silver2014deterministic}
David Silver, Guy Lever, Nicolas Heess, Thomas Degris, Daan Wierstra, and Martin Riedmiller.
\newblock Deterministic policy gradient algorithms.
\newblock In \emph{International Conference On Machine Learning}, pp.\  387--395. Proceedings of Machine Learning Research, 2014.

\bibitem[Stanojev et~al.(2023)Stanojev, Mitridati, Di~Prata, and Hug]{stanojev2023safe}
Ognjen Stanojev, Lesia Mitridati, Riccardo de~Nardis Di~Prata, and Gabriela Hug.
\newblock Safe reinforcement learning for strategic bidding of virtual power plants in day-ahead markets.
\newblock In \emph{IEEE International Conference on Communications, Control, and Computing Technologies for Smart Grids}, pp.\  1--7, 2023.

\bibitem[Stolz et~al.(2024)Stolz, Krasowski, Thumm, Eichelbeck, Gassert, and Althoff]{stolz2024excluding}
Roland Stolz, Hanna Krasowski, Jakob Thumm, Michael Eichelbeck, Philipp Gassert, and Matthias Althoff.
\newblock Excluding the irrelevant: Focusing reinforcement learning through continuous action masking.
\newblock \emph{Advances in Neural Information Processing Systems}, 37:\penalty0 95067--95094, 2024.

\bibitem[Suh et~al.(2022)Suh, Simchowitz, Zhang, and Tedrake]{suh2022differentiable}
Hyung~Ju Suh, Max Simchowitz, Kaiqing Zhang, and Russ Tedrake.
\newblock Do differentiable simulators give better policy gradients?
\newblock In \emph{International Conference on Machine Learning}, pp.\  20668--20696. Proceedings of Machine Learning Research, 2022.

\bibitem[Sutton \& Barto(2018)Sutton and Barto]{sutton2018reinforcement}
Richard~S Sutton and Andrew~G Barto.
\newblock \emph{Reinforcement learning: An introduction}.
\newblock MIT press, 2018.

\bibitem[Tabas \& Zhang(2022)Tabas and Zhang]{tabas2022computationally}
Daniel Tabas and Baosen Zhang.
\newblock Computationally efficient safe reinforcement learning for power systems.
\newblock In \emph{American Control Conference}, pp.\  3303--3310, 2022.

\bibitem[Van~Hasselt(2012)]{van2012reinforcement}
Hado Van~Hasselt.
\newblock Reinforcement learning in continuous state and action spaces.
\newblock In \emph{Reinforcement Learning: State-of-the-Art}, pp.\  207--251. Springer, 2012.

\bibitem[Wabersich et~al.(2023)Wabersich, Taylor, Choi, Sreenath, Tomlin, Ames, and Zeilinger]{Wabersich2023}
Kim~P. Wabersich, Andrew~J. Taylor, Jason~J. Choi, Koushil Sreenath, Claire~J. Tomlin, Aaron~D. Ames, and Melanie~N. Zeilinger.
\newblock Data-driven safety filters: {Hamilton-Jacobi} reachability, control barrier functions, and predictive methods for uncertain systems.
\newblock \emph{IEEE Control Systems Magazine}, 43\penalty0 (5):\penalty0 137--177, 2023.

\bibitem[Wabersich \& Zeilinger(2021)Wabersich and Zeilinger]{wabersich2021predictive}
Kim~Peter Wabersich and Melanie~N Zeilinger.
\newblock A predictive safety filter for learning-based control of constrained nonlinear dynamical systems.
\newblock \emph{Automatica}, 129, 2021.

\bibitem[Walter et~al.(2025)Walter, Markgraf, K{\"u}lz, and Althoff]{walter2025leveraging}
Tim Walter, Hannah Markgraf, Jonathan K{\"u}lz, and Matthias Althoff.
\newblock Leveraging analytic gradients in provably safe reinforcement learning.
\newblock \emph{IEEE Open Journal of Control Systems}, 4:\penalty0 463--481, 2025.

\bibitem[Wang(2022)]{wang2022ensuring}
Xiao Wang.
\newblock Ensuring safety of learning-based motion planners using control barrier functions.
\newblock \emph{IEEE Robotics and Automation Letters}, 7\penalty0 (2):\penalty0 4773--4780, 2022.

\end{thebibliography}
\bibliographystyle{tmlr}

\appendix
\section{Appendix}
\subsection{Action Projection Using Zonotopes}\label{app:zonotopes}

One option for defining safe action sets is to consider control invariant sets as safe state sets $\sX^\varphi$, where $\vx_t \in \sX^\varphi$ ensures that there exists an admissible action $\vu_t \in \sU$ such that \eqref{eq:safety_specs} can be satisfied for all times. The constraints in \eqref{eq:projection_constraints} can be defined using $\sC \subseteq \sX^\varphi$, where $\sC$ is the set of all reachable states at the next time step under the system dynamics and bounded noise $\vw_t \in \sW$. The precise formulation of these constraints depends on the chosen set representations for approximating reachable and safe sets. In this work, we adopt zonotopes to enable efficient computation. Consequently, we have to replace \eqref{eq:projection_constraints} with the necessary constraints for verifying zonotope-in-zonotope containment. Consider a zonotope $\mathcal{Z} \subset \mathbb{R}^{n_z}$ in generator representation that is given by
\begin{equation}\label{eq:zono}
    \mathcal{Z} = \{\vz \in \mathbb{R}^{n_z}: \vz = \vc+\mB \boldsymbol\beta, |\boldsymbol\beta| \leq 1 \}
\end{equation}
with the center $\vc \in \mathbb{R}^{n_z}$ and the generator matrix $\mB \in \mathbb{R}^{n_z \times \eta(\mathcal{Z})}$, where $\eta(\mathcal{Z}) \in \mathbb{N}_0$ denotes the number of generators of $\mathcal{Z}$. The inequality in \eqref{eq:zono} has to be applied elementwise. A more compact notation is given by $\mathcal{Z} = \langle \vc, \mB \rangle_\mathcal{Z}$.
Consider two zonotopes $\mathcal{Z}_1 = \langle \vc_1, \mB_1 \rangle_\mathcal{Z}$,  $\mathcal{Z}_2 = \langle \vc_2, \mB_2 \rangle_\mathcal{Z}$. Then, $\mathcal{Z}_1$ is contained in $\mathcal{Z}_2$, i.e., $\mathcal{Z}_1 \subseteq \mathcal{Z}_2$, if there exist $\boldsymbol\Gamma~\in~\mathbb{R}^{\eta(\mathcal{Z}_2) \times \eta(\mathcal{Z}_1)}$, $\boldsymbol\omega~\in~\mathbb{R}^{\eta(\mathcal{Z}_2)}$ such that \citep[Lemma 1]{ghasemi2020compositional}
\begin{subequations}
\begin{align}
    \mB_1 &= \mB_2 \boldsymbol\Gamma, \\
    \vc_2 - \vc_1 &= \mB_2 \boldsymbol\omega, \\
    \| \begin{bmatrix}
        \boldsymbol\Gamma & \boldsymbol\omega
    \end{bmatrix}\|_\infty &\leq \mathbf{1}.
\end{align}
\label{eq:zono_in_zono}
\end{subequations}
The resulting optimization problem involves both equality and inequality constraints, which is an important consideration when analyzing the sensitivity of the solution, as discussed in appendix \ref{app:diff_safeguard}.

If the safe action set is given directly as a zonotope $\sU^\varphi_\vx = \langle \vc_u, \mB_u \rangle_\mathcal{Z}$, we only have to verify that $\tilde{\vu} \in \sU^\varphi_\vx$. This can be achieved by replacing \eqref{eq:projection_constraints} with \citep{kulmburg2021co}
\begin{align}
    \mB_u \boldsymbol\nu &= \tilde{\vu} - \vc_u \\
    \| \boldsymbol\nu \|_\infty &\leq \mathbf{1}.
\end{align}

\subsection{Differentiating the Safeguard Using the Implicit Function Theorem}\label{app:diff_safeguard}
If the projection safeguard $\Phi$ is integrated into the policy as shown in figure \ref{fig:p2}, we require the sensitivity of its output (the safe action) with respect to its input (the unsafe action) for the backward pass of the policy optimization. 
To obtain the sensitivity of $\vu^\varphi = \Phi(\vx,\vu)$ with respect to $\vu$, let us first consider the Lagrange function associated with the projection problem \ref{eq:projection_problem},
\begin{align}
\mathcal{L}(\vx, \vu, \tilde{\vu}, \boldsymbol\kappa, \boldsymbol\upsilon) &= \Omega(\vx, \vu, \tilde{\vu}) + \boldsymbol\kappa^T D(\vx, \vu, \tilde{\vu}) + \boldsymbol\upsilon^T H(\vx, \vu, \tilde{\vu}) \nonumber \,,
\end{align}
where $\Omega$ is the safeguard objective, and the equality and inequality constraints resulting from the zonotope containment problem in \eqref{eq:zono_in_zono} are represented with $D$ and $H$. The corresponding dual variables are $\boldsymbol\kappa$ and $\boldsymbol\upsilon$, respectively.
Let us further consider the \ac{kkt} conditions
\begin{align}
    \varepsilon(\vx, \vu, \tilde{\vu}, \boldsymbol\kappa, \boldsymbol\upsilon) = \begin{bmatrix}
		\nabla_{\tilde{\vu}}\mathcal{L} (\vx, \vu, \tilde{\vu}, \boldsymbol\kappa, \boldsymbol\upsilon) \\
		D(\vx, \vu, \tilde{\vu}) \\
		\boldsymbol\upsilon^T H(\vx, \vu, \tilde{\vu})
	\end{bmatrix} \,.
\end{align} 
At the \ac{kkt} point, we have
\begin{align*}
    \varepsilon(\vx, \vu, \tilde{\vu}, \boldsymbol\kappa, \boldsymbol\upsilon)|_{\tilde{\vu}=\vu^\varphi, \boldsymbol\kappa=\boldsymbol\kappa^\star, \boldsymbol\upsilon=\boldsymbol\upsilon^\star} = \boldsymbol{0},
\end{align*}
where $\{\vu^{\varphi}, \boldsymbol\kappa^\star, \boldsymbol\upsilon^\star\}$ are the primal-dual solution of the safeguard. 
Then, with the implicit function theorem \citep{krantz2002implicit}, the sensitivity of the safeguard with respect to $\vu$ is 
\begin{align}
\nabla_\vu \Phi (\vx, \vu) = \nabla_{\vu^\varphi} \boldsymbol{\varepsilon}(\vx, \vu, \tilde{\vu}, \boldsymbol\kappa, \boldsymbol\upsilon)^{-1} \nabla_{\vu} \boldsymbol{\varepsilon}(\vx, \vu, \tilde{\vu}, \boldsymbol\kappa, \boldsymbol\upsilon)\ |_{\tilde{\vu}=\vu^\varphi, \boldsymbol\kappa=\boldsymbol\kappa^\star, \boldsymbol\upsilon=\boldsymbol\upsilon^\star} \,.
\end{align}

\subsection{Proof of Theorem 1: Equivalence of Optimal Value Functions}\label{app:proof_theorem_opt_value_functions}

To prove theorem \ref{theorem:eq_opt_val_functions}, let us first recall the definition of an optimal policy. A policy is called optimal if its expected return is greater than or equal to that of all other policies in all states. Therefore, all optimal policies satisfy the same optimal value function \citep{sutton2018reinforcement} 
\begin{equation}\label{eq:def_opt_value_function}
    v^*(\vx) = v_{\pi^\star}(\vx) \geq v_\pi(\vx) \quad \forall \vx \in \tilde{\sX}, \, \pi \in \Pi.
\end{equation}
If we can show that \ac{serl} and \ac{sprl} share the same optimal value function, this implies that any optimal policy in either framework will yield the same expected return. 
Note that policies in both frameworks, $\pi$ and $\pip$, are parameterized by $\boldsymbol\theta$, though this dependence is omitted for readability. 

To demonstrate the equivalence of the optimal value functions, we begin by comparing the conditional probability densities of \ac{serl} -- namely, the state transition density $p_x^\text{SE}$ and the reward density $p_r^\text{SE}$ -- with those of \ac{sprl}, denoted $p_x^\text{SP}$ and $p_r^\text{SP}$. 
At the core of both approaches, we have a sequence of mappings,
\begin{equation}
    \vx \xrightarrow{\pi} \vu \xrightarrow{\Phi} \vu^\varphi \xrightarrow{p^{\text{SP}}} (\vx', r \,),\nonumber
\end{equation}
where each arrow indicates either sampling from a conditional distribution or applying the deterministic transformation $\Phi$.
However, the mappings are grouped differently in \ac{serl} and \ac{sprl}, resulting in the sequences
\begin{align*}
    \text{\ac{sprl}}:\, \vx &\xrightarrow{\pip} \vu^\varphi \xrightarrow{p^{\text{SP}}} (\vx', r) \,, \\
    \text{\ac{serl}}:\, \vx &\xrightarrow{\pi} \vu \xrightarrow{p^{\text{SE}}} (\vx', r) \,. 
\end{align*}
Thus, the Markov processes resulting from policies $\pi$ and $\pip$ with the same $\boldsymbol\theta$ would have the same conditional probability densities, 
\begin{align}
    p^{\text{SE}}_x(\vx'\,|\, \vx) &= p^{\text{SP}}_x(\vx'\,|\, \vx) \,, \label{eq:lemmaeq1}\\
    p^{\text{SE}}_r(r \,|\, \vx) &= p^{\text{SP}}_r(r\,|\, \vx) \,. \label{eq:lemmaeq2}
\end{align}

Now, we can prove theorem \ref{theorem:eq_opt_val_functions}.

\begin{proof}
In \ac{serl}, for a certain policy $\pi$, the state value function (\eqref{eq:v1_bellman}) is given as
\begin{align}
    v_\pi^\text{SE}(\vx) &= \E_{\rvu_t \sim\pi,\, \rvx_t \sim p_x^\text{SE}} \bigg[ \rg_t \bigg| \rvx_0=\vx \bigg] &\\
    &\overset{(\ref{eq:return})}{=} \E_{\rvu_t \sim \pi,\, \rvx_t \sim p_x^\text{SE}} \left[\sum_{k=0}^{\infty} \gamma^k \rr_{t+k+1} \bigg| \rvx_0=\vx \right] &\\
    &\overset{(a)}{=} \sum_{k=0}^\infty \gamma^k \int_{\tilde{\sX}} ... \int_{\tilde{\sX}} \int_\sU ... \int_\sU \int_\sR r \, p^\text{SE}_r(r \, | \,\vx_k, \vu_k)  \pi(\vu_k \,|\, \vx_k)  &\nonumber \\
    &\qquad \qquad \left[ \prod_{i=0}^{k-1}  p^\text{SE}_x(\vx_{i+1} \, | \, \vx_i, \vu_i) \pi(\vu_i \,|\, \vx_i)\right] dr\, \vdu_0 ... \vdu_k\, \vdx_1 ... \vdx_{k} &\\
    &\overset{\text{(LTT)}}{=} \sum_{k=0}^\infty \gamma^k \int_{\tilde{\sX}} ... \int_{\tilde{\sX}} \int_\sR r \, p^{\text{SE}}_r(r\,|\,\vx_k) \left[ \prod_{i=0}^{k-1} p^{\text{SE}}_x(\vx_{i+1}\,|\,\vx_i)\right] dr\, \vdx_1 ... \vdx_{k}, &\label{eq:vserl1}
\end{align}
where LTT refers to the law of total probability \citep[theorem B.70]{schervish1995theory}. Here, $(a)$ subsumes the linearity of expectation as well as the chain rule of probability with the Markov property. A similar derivation can be found in \citet[appendix A.2]{bertsekas2018dynamic}.  

Similarly, for the corresponding safe policy $\pi^\perp$, we have
\begin{align}
    v_{\pip}^\text{SP}(\vx) &= \E_{\rvu_t^\varphi\sim\pi^\perp,\, \rvx_t \sim p_x^\text{SP}} \left[\rg_t \bigg| \rvx_0=\vx \right] &\\
    &\overset{(\ref{eq:return})}{=} \E_{\rvu_t^\varphi\sim\pi^\perp,\, \rvx_t \sim p_x^\text{SP}} \left[\sum_{k=0}^{\infty} \gamma^k \rr_{t+k+1} \bigg| \rvx_0=\vx \right] &\\
    &\overset{(a)}{=} \sum_{k=0}^\infty \gamma^k \int_{\tilde{\sX}} ... \int_{\tilde{\sX}} \int_{\sU^\varphi} ... \int_{\sU^\varphi} \int_\sR r \, p^\text{SP}_r(r \, | \,\vx_k, \vu^\varphi_k) \pip(\vu^\varphi_k \,|\, \vx_k) &\nonumber \\
    &\qquad \qquad \left[ \prod_{i=0}^{k-1} \pip(\vu^\varphi_i \,|\, \vx_i) p^\text{SP}_x(\vx_{i+1} \, | \, \vx_i, \vu^\varphi_i)\right] dr\, \vdu^\varphi_0 ... \vdu^\varphi_k\, \vdx_1 ... \vdx_{k} &\\
    &\overset{\text{(LTT)}}{=} \sum_{k=0}^\infty \gamma^k \int_{\tilde{\sX}} ... \int_{\tilde{\sX}} \int_\sR r \, p^{\text{SP}}_r(r\,|\,\vx_k) \left[ \prod_{i=0}^{k-1} p^{\text{SP}}_x(\vx_{i+1}\,|\,\vx_i)\right] dr\, \vdx_1 ... \vdx_{k}. &\label{eq:vsprl1}
\end{align}
Thus, with \eqref{eq:lemmaeq1} and \eqref{eq:lemmaeq2}, for a policy $\pi \in \Pi$, there exists a safe policy $\pip \in \Pi^\perp$ such that,
\begin{equation}
    v_\pi^\text{SE}(\vx) = v_{\pip}^\text{SP}(\vx) \quad \forall \, \vx \in \tilde{\sX}\,.\label{eq:th1_eq1}
\end{equation}

Consider an optimal policy $\pi^\star$ for \ac{serl}, then the corresponding value function is the optimal value function in \ac{serl}, i.e. $v_{\pi^\star}^{\text{SE}} (\vx) = v^{\text{SE}^\star} (\vx), \forall \vx \in \tilde{\sX}$. Using \eqref{eq:th1_eq1}, there exists a safe policy $\Bar{\pi}^\perp \in \Pi^\perp$ such that $v_{\pi^\star}^{\text{SE}} (\vx) = v_{\Bar{\pi}^\perp}^\text{SP}(\vx), \forall  \vx \in \tilde{\sX}$. However, it is not guaranteed that this safe policy is optimal in the \ac{mdp} $M^\text{SP}$.
Next, we employ a proof by contradiction to establish that the safe policy $\Bar{\pi}^\perp$ and its associated state-value function $v_{\Bar{\pi}^\perp}^\text{SP}$ constitute the optimal policy and optimal state-value function for \ac{sprl}.

Suppose that there exists a safe stochastic policy $\tilde{\pi}^\perp \in \Pi^\perp$ achieving strictly better performance,
\[
    v_{\tilde{\pi}^\perp}^{\text{SP}}(\vx) > v_{\bar{\pi}^\perp}^{\text{SP}}(\vx), \quad \text{for some } \vx \in \tilde{\sX}.
\]

Since $\tilde{\pi}^\perp(\cdot\,|\, \vx)$ is supported on $\sU^\varphi \subseteq \sU$, we can construct a corresponding SE-RL policy $\tilde{\pi}(\cdot\,|\, \vx)$ whose pushforward through $\Phi$ equals $\tilde{\pi}^\perp(\cdot\,|\, \vx)$. Formally, for each $\vx$:
\[
\Phi_\# \tilde{\pi}(\cdot\,|\, \vx) = \tilde{\pi}^\perp(\cdot\,|\, \vx),
\]
where $\Phi_\#$ denotes the pushforward of the probability measure under $\Phi$.  

\emph{Intuition:} for each safe action $\vu^\varphi$ sampled from $\tilde{\pi}^\perp(\cdot\,|\, \vx)$, select a preimage $\vu \in \Phi^{-1}(\vu^\varphi)$ according to any probability distribution over that set. This is always possible because $\Phi$ is \emph{surjective}, ensuring every $\vu^\varphi \in \sU^\varphi$ has at least one preimage in $\sU$. 

Using \eqref{eq:th1_eq1}, the state-transition and reward distributions under $\tilde{\pi}$ in SE-RL match those under $\tilde{\pi}^\perp$ in SP-RL. Hence,
\[
    v_{\tilde{\pi}}^{\text{SE}}(\vx) = v_{\tilde{\pi}^\perp}^{\text{SP}}(\vx) > v_{\bar{\pi}^\perp}^{\text{SP}}(\vx) = v^{\text{SE}^\star}(\vx),
\]
contradicting the optimality of $\pi^\star$.  

Therefore, no such $\tilde{\pi}^\perp$ exists, and $\bar{\pi}^\perp$ is also optimal in SP-RL. Consequently, the optimal value functions coincide,
\[
    v^{\text{SE}^\star}(\vx) = v^{\text{SP}^\star}(\vx), \quad \forall \vx \in \tilde{\sX}.
\]

\end{proof}

\subsection{Proof of Lemma 1: Equivalence of SE-RL and SP-RL for Stochastic Policies and GAE}\label{app:proof_lemma_1}
\begin{figure*}[t]
    \centering
    \begin{subfigure}[t]{0.32\textwidth}
        \includesvg[width=\textwidth]{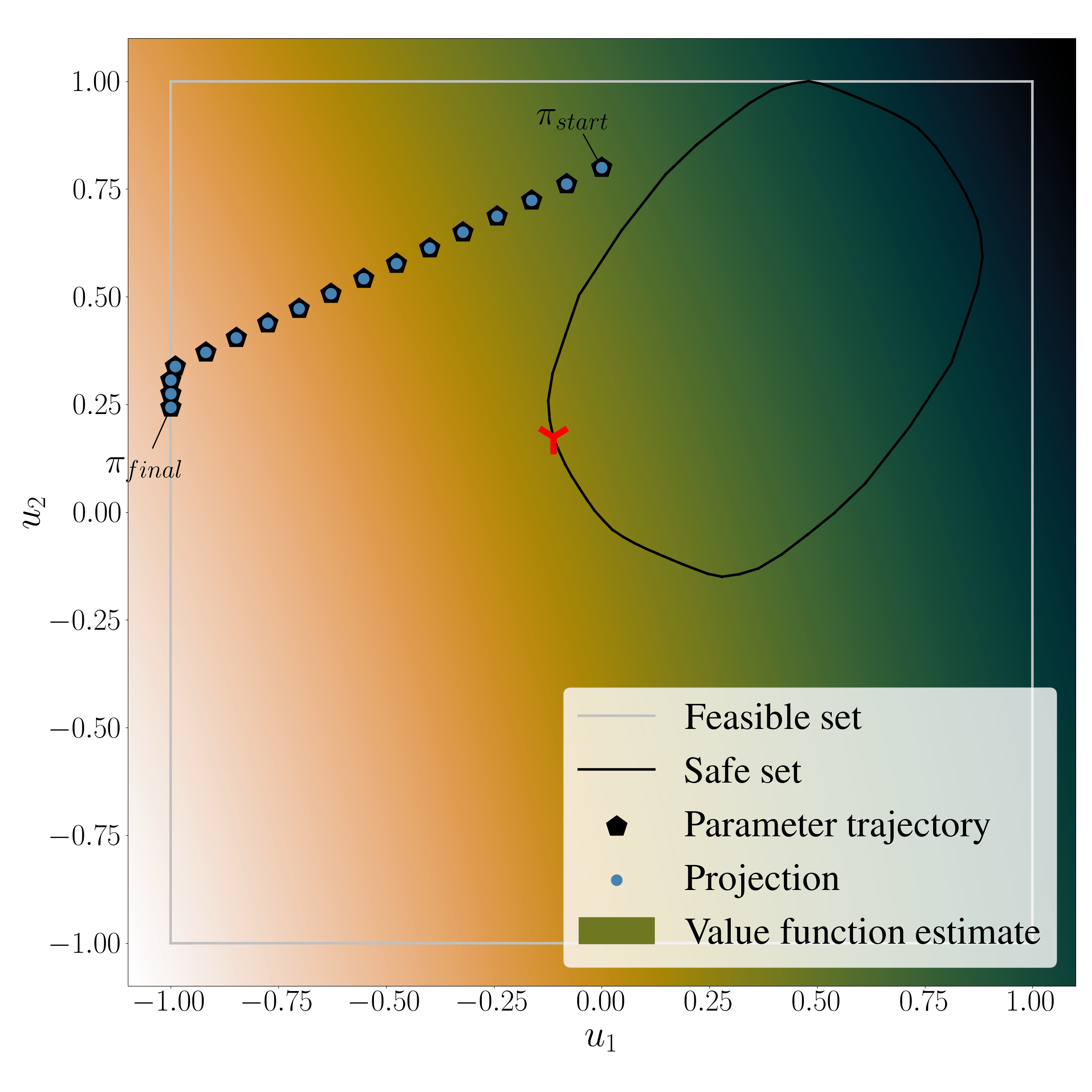}
        \caption{Without safeguarding, the critic approximates the true objective function, showing that the optimal action would be unsafe. Note that we clip the policy to the feasible action set during policy improvement.}
        \label{fig:min_ex_seeker_unsafe}
    \end{subfigure}
    \begin{subfigure}[t]{0.32\textwidth}
        \includesvg[width=\textwidth]{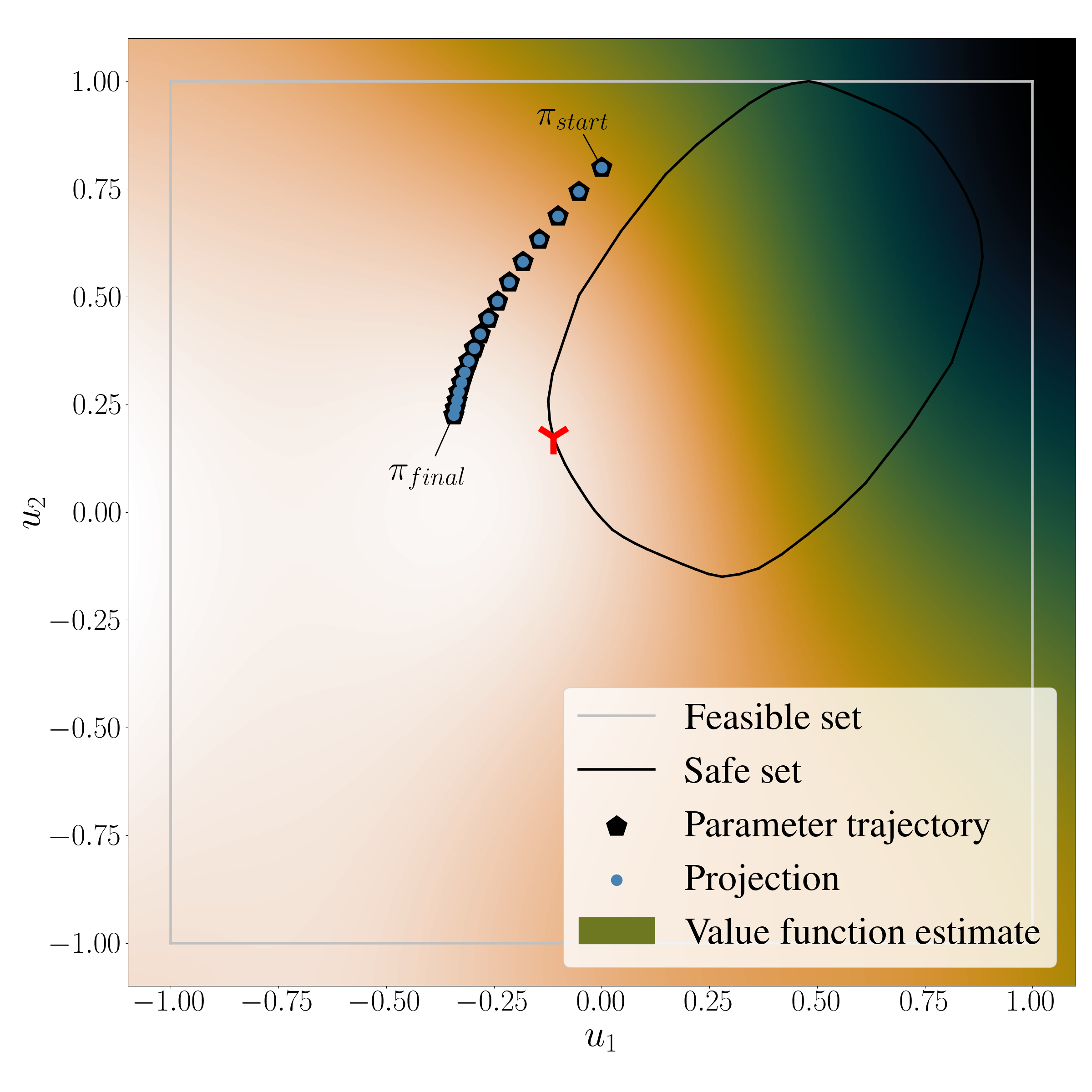}
        \caption{\ac{serl}: Policy converges to an unsafe action that lies in the direction normal to the safe optimal action and the safe set boundary.}
        \label{fig:min_ex_seeker_serl}
    \end{subfigure}    
    \begin{subfigure}[t]{0.32\textwidth}
        \includesvg[width=\textwidth]{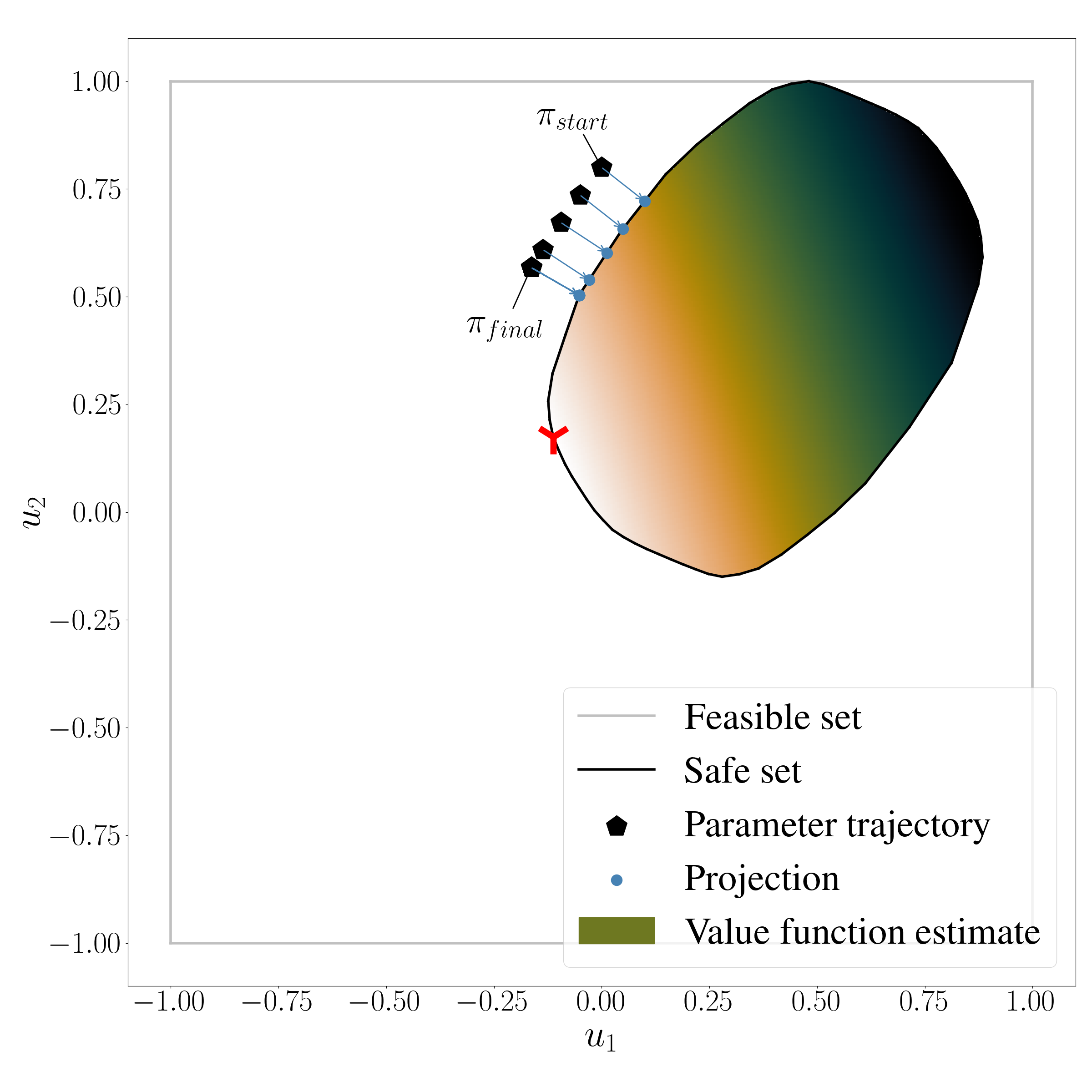}
        \caption{\ac{sprl}: Policy with differentiable projection safeguard does not improve in the direction normal to the projection and eventually gets stuck on a vertex of the safe action set.}
        \label{fig:min_ex_seeker_sprl}
    \end{subfigure}
    \caption{Effect of action aliasing on \ac{serl} and \ac{sprl} algorithms using deterministic policies.  We illustrate the policy improvement step for a given state $\vx$ in the navigation task. The deterministic policy $\pi(\vx)$ is updated for 50 steps using a loss function based on the learned state-action value function $q(\vx, \vu)$ in the \ac{serl} case and $q(\vx, \vu^\varphi)$ in the \ac{sprl} case.} 
    \label{fig:min_example_seeker}
\end{figure*}
\begin{figure*}[t]
    \centering
    \begin{subfigure}[t]{0.32\textwidth}
        \includesvg[width=\textwidth]{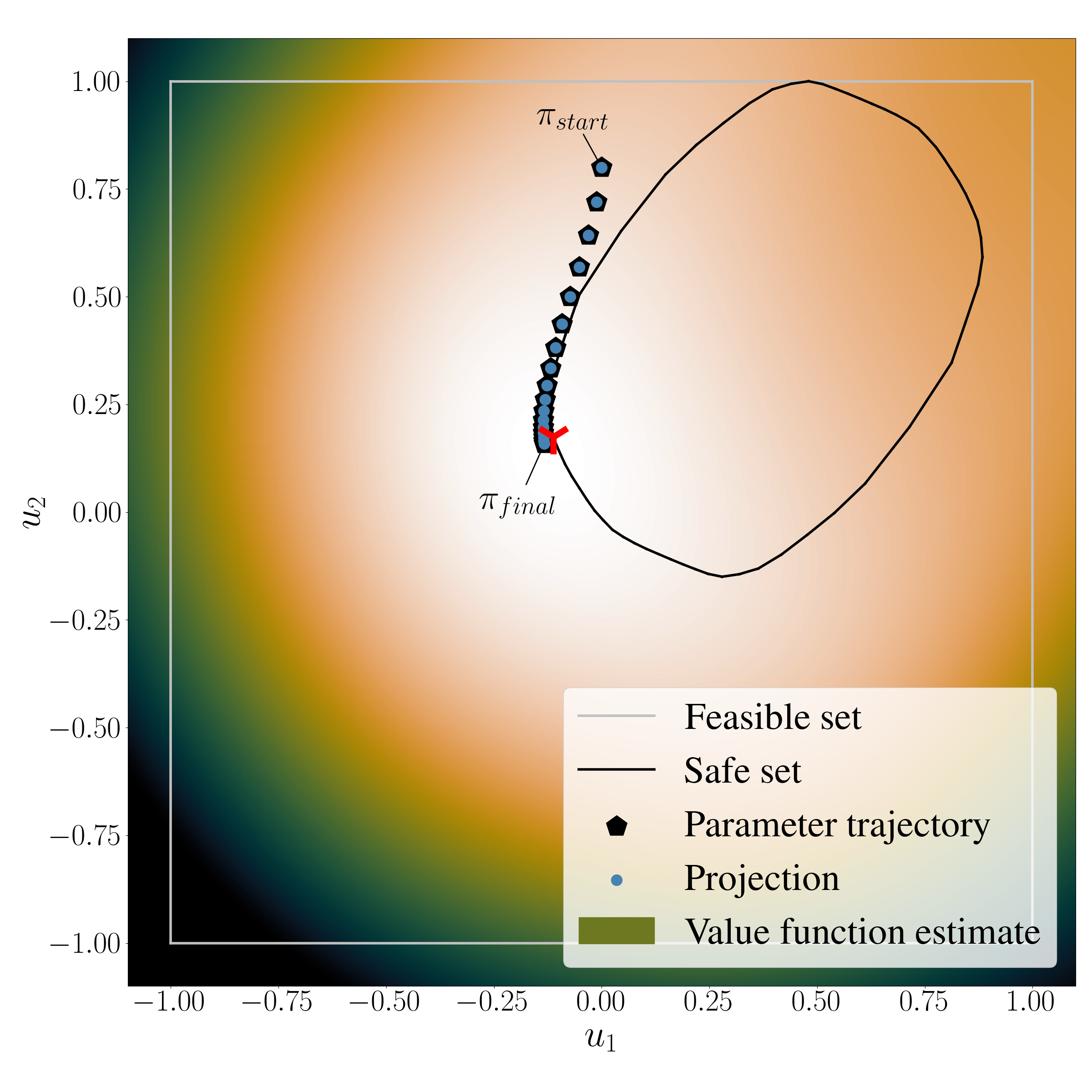}
        \caption{\ac{serl}: Adding a penalty to the reward addresses the flat-lining critic problem and improves convergence toward the optimal safe action.}
        \label{fig:min_ex_seeker_serl_pen}
    \end{subfigure}
    \begin{subfigure}[t]{0.32\textwidth}
        \includesvg[width=\textwidth]{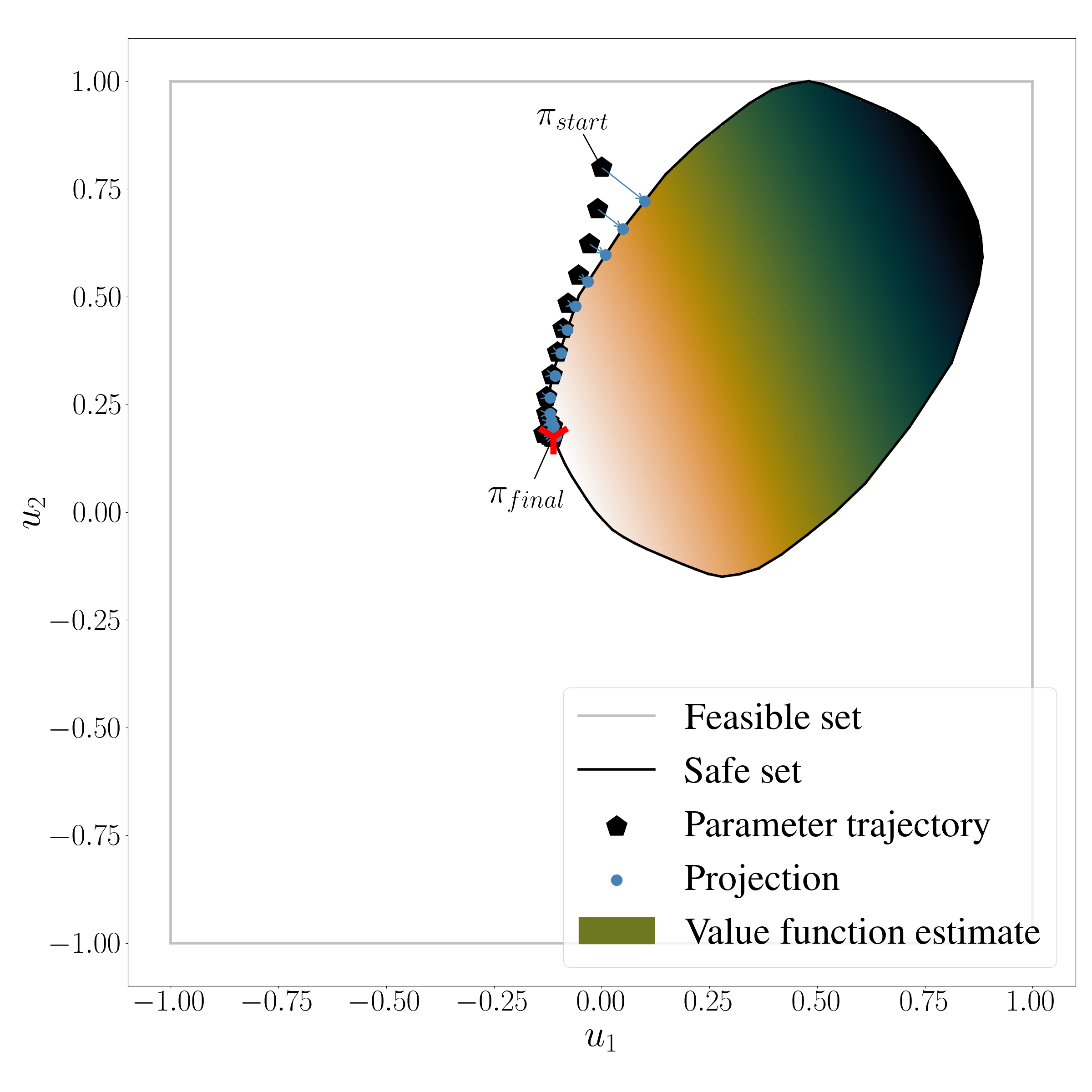}
        \caption{\ac{sprl}: An additional policy loss term that penalizes the distance between unsafe and safe action improves convergence to the optimal safe action. However, if the optimal safe action would lie in the interior of the safe action set, it might not be reached.}
        \label{fig:min_ex_seeker_sprl_loss}
    \end{subfigure}    
    \begin{subfigure}[t]{0.32\textwidth}
        \includesvg[width=\textwidth]{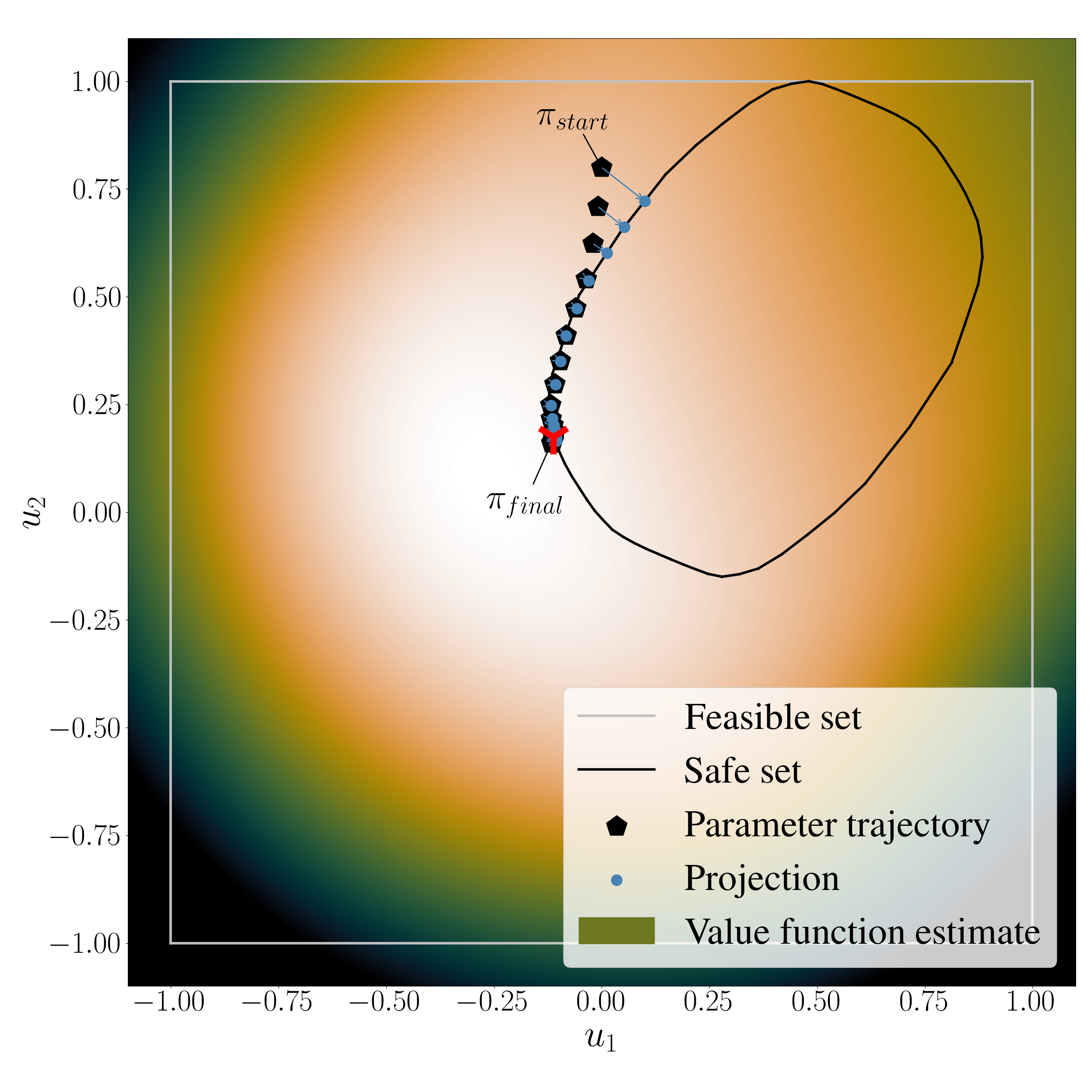}
        \caption{\ac{sprl}: Learning an additional penalty critic is very similar to adding a penalty to the reward in \ac{serl}. Note that as the penalty critic is conditioned on unsafe actions, we display the objective over the entire action space.}
        \label{fig:min_ex_seeker_sprl_critic}
    \end{subfigure}
    \caption{Effect of improvement strategies when using a differentiable safeguard during policy updates for a given state $\vx$ in the navigation task. The deterministic policy $\pi_{\boldsymbol\theta}(\vx)$ is updated for 50 steps using a loss function based on the learned state-action value function $q(\vx, \vu)$ in the \ac{serl} case and $q(\vx, \vu^\varphi)$ in the \ac{sprl} case}. 
    \label{fig:min_example_improved_seeker}
\end{figure*}
\begin{proof}
With $\Psi^\text{SP}_{\pip}(\vx,\vu^\varphi)=\hat{a}^\text{GAE}(\vx, \vu^\varphi)$ and $\Psi^\text{SE}_{\pip}(\vx,\vu)=\hat{a}^\text{GAE}(\vx, \vu)$, we can rewrite the policy gradient estimates for \ac{sprl} for stochastic policies as
    \begin{align*}
        \nabla_{\boldsymbol\theta} J(\pip_{\boldsymbol\theta}) &= \E_{\pip}\left[\Psi^\text{SP}_{\pip}(\vx,\vu^\varphi) \nabla_{\boldsymbol\theta} \log \pi_{\boldsymbol\theta}(\vu\,|\,\vx)\right] & \\
        &= \E_{\pip}\left[\hat{a}^\text{GAE}_{\pip}(\vx,\vu^\varphi) \nabla_{\boldsymbol\theta} \log \pi_{\boldsymbol\theta}(\vu\,|\,\vx)\right] & \\
        &\overset{(\ref{eq:gae})}{=} \E_{\pip}\left[\sum_{l=0}^\infty (\gamma \lambda)^l \delta^v_{t+l} \nabla_{\boldsymbol\theta} \log \pi_{\boldsymbol\theta}(\vu\,|\,\vx)\right] &  \\
        &\overset{(\ref{eq:td_residual_v})}{=} \E_{\pip}\left[\sum_{l=0}^\infty (\gamma \lambda)^l (r_{t+l} + \gamma v_{\pip}^\text{SP}(\vx_{t+l+1}) - v_{\pip}^\text{SP}(\vx_{t+l})) \nabla_{\boldsymbol\theta} \log \pi_{\boldsymbol\theta}(\vu\,|\,\vx)\right] & \\
    \end{align*}
and, similarly,
    \begin{equation*}
        \nabla_{\boldsymbol\theta} J(\pi_{\boldsymbol\theta}) = \E_{\pi}\left[\sum_{l=0}^\infty (\gamma \lambda)^l (r_{t+l} + \gamma v_{\pi}^\text{SE}(\vx_{t+l+1}) - v_{\pi}^\text{SE}(\vx_{t+l})) \nabla_{\boldsymbol\theta} \log \pi_{\boldsymbol\theta}(\vu\,|\,\vx)\right].
    \end{equation*}
Using the same initial parameters $\boldsymbol\phi = \boldsymbol\phi^\text{in}$ and ${\boldsymbol\theta} = {\boldsymbol\theta}^\text{in}$, both policy gradient estimates are the same for the first policy update. Similarly, as the value functions $v_{\pi}^\text{SE}$ and $v_{\pip}^\text{SP}$ only depend on the state, the gradient $\nabla_{\boldsymbol\phi} L_{\boldsymbol\phi}$ for the loss defined in \eqref{eq:critic_loss} is also equivalent. Consequently, if all other algorithm hyperparameters are the same for both \ac{serl} and \ac{sprl}, all subsequent parameter updates will also be the same.
\end{proof}

\subsection{Proof of Lemma 3: Flat-lining Critic in SE-RL}\label{app:proof_lemma_3}
\begin{proof}
    Any action $\vu^e \in \sU^e(\vu^b)$ is projected to $\vu^b$ and thus transitions to the same next $\vx'$ and yields the same reward $r^b \sim p_r(r \, | \, \vx, \vu^b)$. Since the state-action value function satisfies \citep{sutton2018reinforcement}
    \begin{equation*}
        q_\pi(\vx,\vu) = \E\left[r + \gamma v_\pi(\vx') \,|\, \rvx_t=\vx, \rvu_t=\vu\right],
    \end{equation*}
    we obtain
    \begin{equation*}
        q_\pi(x,\vu^e) = q_\pi(\vx,\vu^b) \quad \forall \vu^e \in \sU^e,  
    \end{equation*}
and, consequently, 
\begin{equation*}
        a_\pi(\vx,\vu^e) = a_\pi(\vx,\vu^b) \quad \forall \vu^e \in \sU^e.  
    \end{equation*}
Furthermore, for the \ac{gae} according to \eqref{eq:gae}, we receive
\begin{equation*}
        \hat{a}^{\text{GAE}}_\pi(\vx,\vu^e) = \hat{a}^{\text{GAE}}_\pi(\vx,\vu^b) \quad \forall \vu^e \in \sU^e.  
    \end{equation*}
\end{proof}

\subsection{Action Aliasing Visualization for Seeker Task}\label{app:min_example_seeker}
We provide a second minimal example to visualize the impact of action aliasing in deterministic policies. It is based on the seeker navigation task described in appendix \ref{app:benchmark_problems}. All other settings are the same as listed in section \ref{sec:action_aliasing}. The main difference to the previously described example is that task performance and safety are not aligned in the navigation task, as one obstacle will always be situated between the initial and the goal position. Therefore, actions that are optimal with respect to the task performance are usually unsafe, and the optimal safe action lies on the boundary of the safe action set. This is visualized in figure \ref{fig:min_example_seeker}. Due to these different task characteristics, the impact of the different improvement strategies becomes more similar as shown in figure \ref{fig:min_example_improved_seeker}.

\subsection{Details of Penalty Critic Implementation}\label{app:penalty_critic_details}
\begin{figure}
    \centering
    \includesvg[width=\linewidth]{figures/penalty_critic_analysis.svg}
    \caption{Comparison of the penalty critic loss and the number of safeguard interventions during training for different scaling factors $w$.}
    \label{fig:penc_stability}
\end{figure}
The penalty critic is a separate neural network with the same architecture as the task critic. In contrast to the latter, the penalty critic is conditioned on the unsafe action $\vu$, not the safe action. Its parameters $\varsigma$ are trained using the mean-squared error loss in \eqref{eq:critic_loss}, where the target $y$ is computed as
\begin{equation}
    y_t = \underbrace{w\|\vu_t - \vu^\varphi_t\|^2_2}_{h_t} + \gamma q^\text{pen}_{\pi, \tilde{\boldsymbol\varsigma}}(\vx_{t+1},\vu_{t+1}).
\end{equation}
Consequently, the scaling factor $w$ has a strong impact on the learning stability of the penalty critic. Figure \ref{fig:penc_stability} shows that while for the pendulum and the EMS all penalty factors lead to a stable loss progression and reduced safeguard interventions, only higher values for $w$ can achieve this for the quadrotor environment. For the seeker navigation task, where performance and safety are not aligned (see section \ref{sec:benchmarks}), a low penalty factor enables the best convergence behavior.

\subsection{Benchmark Problems}\label{app:benchmark_problems}
\subsubsection{Pendulum Stabilization Task}
Our pendulum environment is closely related to the \textit{OpenAI Gym Pendulum-V0}\footnote{\url{https://gymnasium.farama.org/environments/classic_control/pendulum/}} environment with the difference that we limit the one-dimensional control input to $|u| \leq 8 \,\text{rad}\text{s}^{-1}$. The environment has the state $\vx=\begin{bmatrix}\vartheta, \dot \vartheta\end{bmatrix}^T$ and the dynamics
\begin{equation}
    \dot \vx = \begin{pmatrix}
        \dot \vartheta \\ \frac{g}{\ell} \sin(\vartheta) + \frac{1}{m\ell^2} u
    \end{pmatrix},
\end{equation}
where $g$ is gravity and $m, \ell$ are the mass and the length of the pendulum, respectively. We discretize the dynamics using the explicit Euler method. The desired equilibrium state is $\vx^*=\begin{bmatrix}0, 0\end{bmatrix}^T$. The reward is $r=-(\vartheta^2 + 0.1 \dot \vartheta^2 + 0.001 u^2)$. 

\subsubsection{Quadrotor Stabilization Task}
We use the model of a quadrotor operating in the $x$-$z$-plane that is proposed in \citep{stolz2024excluding}. The system has two independent thrusts $\vu=\begin{bmatrix} u^1, u^2 \end{bmatrix}^T$ bounded by lower and upper limits $\underline{\vu}, \overline{\vu}$, respectively. The state of the system is $\vx=\begin{bmatrix}
    e^x, e^z, \dot e^x, \dot e^z,  \vartheta, \dot \vartheta
\end{bmatrix}^T$, where $e_{x,z}$ are the positions along the $x$- and $z$-axis, respectively. The system dynamics are
\begin{equation}
    \dot \vx = \begin{pmatrix}
        \dot e^x \\ \dot e^z \\ (u^1+u^2) K \sin(\vartheta) \\ -g + (u^1+u^2) K \cos(\vartheta) \\ \dot \vartheta \\ -d_0 \vartheta -d_1 \dot \vartheta + n_0 (-u^1 + u^2)
    \end{pmatrix} + \begin{pmatrix}
        0 \\ 0 \\ w^1 \\ w^2 \\ 0 \\ 0
    \end{pmatrix}, 
\end{equation}
where $d_0, d_1, n_0, K$ are constants and $\vw=\begin{bmatrix}
    w^1, w^2
\end{bmatrix}^T$ are disturbances. The linearized dynamics are obtained using a first-order Taylor expansion around the equilibrium point $\vx^* = \begin{bmatrix}
    0, 1, 0, 0, 0, 0
\end{bmatrix}^T$ and are discretized in time using the explicit Euler method. The disturbances are sampled uniformly from a compact disturbance set $\sW\subset \mathbb{R}^2$. The reward is computed using $r = -1 + \exp \left(-\|\vs - \vs^*\|_2 - \frac{0.01}{2} \|\left[\frac{u^1 - \underline{u}^1}{\overline{u}^1 - \underline{u}^1}, \frac{u^2 - \underline{u}^2}{\overline{u}^2 - \underline{u}^2}\right] \|_1 \right)$, where $\vs = \begin{bmatrix}
    e^x, e^z
\end{bmatrix}$. 

\subsubsection{Seeker Navigation Task}
In this two-dimensional navigation task, a simple massless robot has to find the shortest path to the goal while avoiding a fixed number of obstacles. The environment is configured such that at least one obstacle always lies between the initial position of the seeker and the goal position $\ve^{g}$. The initial position, goal position, and the positions and radii of the obstacles are pseudo-randomly sampled at the beginning of each episode. The state $\vx = [e^{x}, e^{y}, \dot e^{x}, \dot e^{y}]^{T}$ consists of the positions and velocities in the $x$-$y$-plane. The actions $\vu=[\ddot e^{x}, \ddot e^{y}]$ are accelerations in the respective directions and bounded by $|\ddot e^{x}|, |\ddot e^{y}| \leq a^{\max}$ with $a^{\max} = 1\,m/s^2$. We use the simplified system dynamics 
\begin{equation}
    \dot \vx = \begin{pmatrix}
        \dot e^x \\ \dot e^y \\ \ddot e^x \\ \ddot e^y
    \end{pmatrix} + \begin{pmatrix}
        0 \\ 0 \\ w^1 \\ w^2
    \end{pmatrix},
\end{equation}
and discretize them using the explicit Euler method. The disturbances $\vw=\begin{bmatrix}
    w^1, w^2
\end{bmatrix}^T$ are uniformly sampled from a compact disturbance set $\sW \subset \sR^2$. The position of the seeker at time $t$ is $\ve_t$, and the step reward is $r = -1 + \exp(\|\ve_t - \ve_g\|)$. 
In all experiments, we set the number of obstacles to three. 

In each time step, a safe action set $\sU^\varphi_\vx(x)$ is computed to avoid collisions with obstacles and meet the boundaries of the map. This simplifies the projection problem in \eqref{eq:projection_problem} to 
\begin{align}
    \Phi(\vx, \vu): \nonumber \\
    \vu^\varphi= &\argmin_{\tilde{\vu}} \frac{1}{2} \|\tilde{\vu} - \vu \|^2_2    \\
    \text{s.t.} \quad & \tilde{\vu} \in \sU^\varphi_\vx. \\ 
\end{align}

\subsubsection{Energy Management System Optimization Task}
The goal in this task (introduced in \citet{walter2025leveraging}) is to control a battery and a heat pump in a building that has a non-flexible electric base load and a non-controllable PV generator. The state $\vx = [soc, \Theta^\text{in}, \Theta^\text{ret}]^T$ comprises the state of charge of the battery, the indoor temperature, and the return temperature of the floor heating system. The policy of the agent is conditioned on the observation $\vo = [\vx, \Theta^\text{out}_{t:t+H}, p^\text{PV}_{t:t+H}, p^\text{L}_{t:t+H}, \varrho_{t:t+H}]$, which is composed of the current value and $H$ forecasts of the outdoor temperature, the PV power, the inflexible load, and the electricity price. We use $H=5$ such that the observation space has a dimensionality of $23$. The actions $\vu = [p^\text{B}, p^\text{HP}]$ are the continuous power set points for the battery and the heat pump, respectively. The system follows the simplified dynamics
\begin{equation}
    \dot{\vx} = \begin{pmatrix}
        p^\text{B} \\ -c_0 \Theta^\text{in} + c_1 \Theta^\text{ret} \\ c_2 \Theta^\text{in} - c_2 \Theta^\text{ret} + c_3 p^\text{HP}
    \end{pmatrix} + \begin{pmatrix}
        0 \\ c_4 \Theta^\text{out} \\ 0
    \end{pmatrix},
\end{equation}
where the outdoor temperature acts as a bounded noise. 
The computation of the coefficients $c_{0-4}$ is detailed in \citet[equation 2.17]{bianchi2006adaptive}.
Performance is measured through the deviation of the indoor temperature from a desired set point $\Theta^\text{set}$ as well as through the electricity costs such that $r = -(p^\text{B} + p^\text{HP} + p^\text{L} - p^\text{PV}) \varrho - (\Theta^\text{in} - \Theta^\text{set})^2$. The state constraints $\sX = [0,10] \times [18, 24] \times [10,100]$ are enforced using reachability analysis. During training, one episode corresponds to one day, and the initial state is randomized.

\subsection{Wall Clock Time Results}\label{app:wall_clock_times}
\begin{table}[h!]
\centering
\caption{Comparison of wall clock time for TD3 (deterministic policy case) over 10,000 training steps, normalized to the vanilla SE-RL approach.}
\label{tab:time_comparison}
\begin{tabular}{@{}lcccccc@{}}
\toprule
          & \multicolumn{1}{l}{\textbf{States}} & \multicolumn{1}{l}{\textbf{SE-RL}} & \multicolumn{1}{l}{\textbf{SE-RL Penalty}} & \multicolumn{1}{l}{\textbf{SP-RL}} & \multicolumn{1}{l}{\textbf{SP-RL PSL}} & \multicolumn{1}{l}{\textbf{SP-RL PenC}} \\ \midrule
\textbf{Pendulum}  & 2                                          & 1.0                       & 0.99                              & 6.04                      & 5.59                          & 5.49                           \\
\textbf{Seeker}   & 2                                          & 1.0                       & 1.0                               & 3.56                      & 3.55                          & 3.71                           \\
\textbf{Quadrotor} & 6                                          & 1.0                       &             1.0                      &          12.01                 &                 10.51              &               9.02                 \\
\textbf{EMS} & 3                                       & 1.0                       &             0.98                     &          12.25                &                 12.38            &               12.51                 \\\bottomrule
\end{tabular}
\end{table}

\newpage
\subsection{Comprehensive Results}\label{app:full_results}

\begin{table}[h]
\centering
\caption{\ac{td3}, pendulum: mean and standard deviation of returns and safeguard interventions for different scaling factors $w$.}
\label{tab:pend_td3}
\begin{tabular}{@{}lccccc@{}}
\toprule
\textbf{} & \textbf{SE-RL} & \textbf{SE-RL Penalty} & \textbf{SP-RL} & \textbf{SP-RL PSL} & \textbf{SP-RL PenC} \\ \midrule
& \multicolumn{5}{c}{$w=0.1$} \\ \midrule
\textbf{Interventions}      & $0.01 \pm 0.20$   & $0.03 \pm 0.24$   & $4.13 \pm 5.10$ & $1.33 \pm 1.35$ & $5.49 \pm 18.23$ \\
\textbf{Returns} & $-11.52 \pm 7.24$ & $-13.23 \pm 8.19$ & $-10.52 \pm 7.41$ & $-8.06 \pm 6.16$ & $-8.71 \pm 6.22$ \\ 
\midrule
& \multicolumn{5}{c}{$w=0.5$} \\ \midrule
\textbf{Interventions}       & $0.01 \pm 0.20$   & $0.0 \pm 0.0$   & $4.13 \pm 5.10$ & $2.51 \pm 4.17$ & $8.90 \pm 18.29$ \\
\textbf{Returns}            & $-11.52 \pm 7.24$ & $-29.34 \pm 19.43$ & $-10.52 \pm 7.41$ & $-8.77 \pm 6.21$ & $-7.60 \pm 5.99$ \\ 
\midrule
& \multicolumn{5}{c}{$w=1.0$} \\ \midrule
\textbf{Interventions}       & $0.01 \pm 0.20$   & $0.07 \pm 0.35$   &  $4.13 \pm 5.10$ & $1.36 \pm 1.46$ & $13.06 \pm 27.32$ \\
\textbf{Returns}           & $-11.52 \pm 7.24$ & $-23.26 \pm 21.19$ & $-10.52 \pm 7.41$ & $-8.33 \pm 6.84$ & $-9.57 \pm 7.08$ \\
\midrule
& \multicolumn{5}{c}{$w=2.0$} \\ \midrule
\textbf{Interventions}       & $0.01 \pm 0.20$   & $0.0 \pm 0.0$   & $4.13 \pm 5.10$ & $18.57 \pm 34.74$ & $16.53 \pm 25.80$ \\
\textbf{Returns}           & $-11.52 \pm 7.24$ & $-45.48 \pm 23.71$ & $-10.52 \pm 7.41$ & $-8.50 \pm 6.30$ & $-7.47 \pm 5.70$ \\ 
\bottomrule
\end{tabular}
\end{table}

\begin{table}[h]
\centering
\caption{\ac{td3}, quadrotor: mean and standard deviation of returns and safeguard interventions for different scaling factors $w$.}
\label{tab:quad_td3}
\begin{tabularx}{\textwidth}{@{}Xccccc@{}}
\toprule
\textbf{} & \textbf{SE-RL} & \textbf{SE-RL Penalty} & \textbf{SP-RL} & \textbf{SP-RL PSL} & \textbf{SP-RL PenC} \\ \midrule
& \multicolumn{5}{c}{$w=0.1$} \\ \midrule
\textbf{Interventions} & $194.23 \pm 14.01$ & $170.13 \pm 48.93$ & $199.71 \pm 0.64$ & $199.04 \pm 1.49$ & $192.60 \pm 23.79$ \\
\textbf{Returns} & $-49.43 \pm 6.76$ & $-49.75 \pm 12.35$ & $-66.34 \pm 16.76$ & $-66.87 \pm 17.41$ & $-62.52 \pm 13.00$ \\
\midrule
& \multicolumn{5}{c}{$w=0.5$}\\ \midrule
\textbf{Interventions} & $194.23 \pm 14.01$ & $64.16 \pm 46.66$ & $199.71 \pm 0.64$ & $170.44 \pm 51.05$ & $145.31 \pm 49.21$ \\
\textbf{Returns} & $-49.43 \pm 6.76$ & $-43.76 \pm 13.68$ & $-66.34 \pm 16.76$ & $-59.11 \pm 18.65$ & $-53.46 \pm 17.47$ \\ 
\midrule
& \multicolumn{5}{c}{$w=1.0$}\\ \midrule
\textbf{Interventions}      & $194.23 \pm 14.01$ & $65.63 \pm 60.98$ & $199.71 \pm 0.64$ & $181.79 \pm 37.02$ & $141.80 \pm 58.84$\\
\textbf{Returns}            & $-49.43 \pm 6.76$ & $-43.33 \pm 16.68$ & $-66.34 \pm 16.76$ & $-49.97 \pm 7.68$       & $-54.22 \pm 16.73$ \\ 
\midrule
& \multicolumn{5}{c}{$w=2.0$}\\ \midrule
\textbf{Interventions} & $194.23 \pm 14.01$ & $45.94 \pm 40.76$ & $199.71 \pm 0.64$ & $192.51 \pm 21.70$ & $134.93 \pm 52.98$\\
\textbf{Returns} & $-49.43 \pm 6.76$ & $-42.81 \pm 11.78$ & $-66.34 \pm 16.76$ & $-49.34 \pm 5.74$ & $-44.34 \pm 10.26$\\
\bottomrule
\end{tabularx}
\end{table}

\begin{table}[h]
\centering
\caption{\ac{td3}, seeker: mean and standard deviation of returns and safeguard interventions for different scaling factors $w$.}
\label{tab:seeker_td3}
\begin{tabular}{@{}lccccc@{}}
\toprule
\textbf{} & \textbf{SE-RL} & \textbf{SE-RL Penalty} & \textbf{SP-RL} & \textbf{SP-RL PSL} & \textbf{SP-RL PenC} \\ \midrule
              & \multicolumn{5}{c}{$w=0.1$}                                     \\ \midrule
\textbf{Interventions} & $83.90 \pm 34.00$ & $85.60 \pm 32.19$  & $81.7 \pm 35.16$  &  $42.24 \pm 43.90$ & $49.90 \pm 44.94$ \\
\textbf{Returns}      & $-34.20 \pm 20.05$ & $-34.16 \pm 20.14$ & $-51.33 \pm 22.97$ & $-26.31 \pm 20.13$ &  $-31.66 \pm 20.43$\\
\midrule
              & \multicolumn{5}{c}{$w=0.5$}                                     \\ \midrule
\textbf{Interventions} & $83.90 \pm 34.00$ & $80.00 \pm 37.10$  & $81.7 \pm 35.16$  & $44.01 \pm 39.53$ & $42.63 \pm 44.65$ \\
\textbf{Returns}     & $-34.20 \pm 20.05$ & $-32.73 \pm 20.70   $  & $-51.33 \pm 22.97$ & $-26.10 \pm 18.19$ & $-31.02 \pm 19.22$ \\ 
\midrule
              & \multicolumn{5}{c}{$w=1.0$}                                     \\ \midrule
\textbf{Interventions} & $83.90 \pm 34.00$ & $93.96 \pm 22.04$ & $81.7 \pm 35.16$   & $43.19 \pm 38.87$ & $46.03 \pm 45.75$ \\
\textbf{Returns} & $-34.20 \pm 20.05$ & $-39.12 \pm 16.59$ & $-51.33 \pm 22.97$ & $-28.17 \pm 17.49$ & $-36.06 \pm 19.26$ \\
\midrule
              & \multicolumn{5}{c}{$w=2.0$}                                     \\ \midrule
\textbf{Interventions} & $83.90 \pm 34.00$ & $97.47 \pm 14.86$  & $81.7 \pm 35.16$   & $40.40 \pm 39.20$ &  $39.31 \pm 2.97$\\
\textbf{Returns} & $-34.20 \pm 20.05$ & $-48.09 \pm 15.86$  & $-51.33 \pm 22.97$ & $-28.90 \pm 19.73$ & $-42.04 \pm 17.69$ \\
\bottomrule
\end{tabular}
\end{table}

\begin{table}[]
\caption{TD3, EMS: mean and standard deviation of returns and safeguard interventions for different scaling factors $w$.}
\label{tab:ems_td3}
\begin{tabular}{@{}lccccc@{}}
\toprule
                       & \textbf{SE-RL}    & \textbf{SE-RL Penalty} & \textbf{SP-RL}     & \textbf{SP-RL PSL} & \textbf{SP-RL PenC} \\ \midrule
\textbf{}              & \multicolumn{5}{c}{$w=0.1$}                                                                                  \\ \midrule
\textbf{Interventions} & $17.19 \pm 7.42$  & $3.1 \pm 4.36$         & $22.99 \pm 0.12$   & $16.23 \pm 4.46$   & $14.11 \pm 7.75$    \\
\textbf{Returns}       & $-16.11 \pm 8.81$ & $-18.71 \pm 8.77$      & $-26.45 \pm 19.25$ & $-21.60 \pm 15.51$ & $-14.59 \pm 6.46$   \\ \midrule
\textbf{}              & \multicolumn{5}{c}{$w=0.5$}                                                                                  \\ \midrule
\textbf{Interventions} & $17.19 \pm 7.42$  & $0.43 \pm 0.79$        & $22.99 \pm 0.12$   & $13.9 \pm 8.78$    & $15.13 \pm 4.55$    \\
\textbf{Returns}       & $-16.11 \pm 8.81$ & $-20.02 \pm 8.65$      & $-26.45 \pm 19.25$ & $-15.63 \pm 7.11$  & $-12.65 \pm 4.59$   \\ \midrule
\textbf{}              & \multicolumn{5}{c}{$w=1.0$}                                                                                  \\ \midrule
\textbf{Interventions} & $17.19 \pm 7.42$  & $0.0 \pm 0.0$          & $22.99 \pm 0.12$   & $15.81 \pm 4.94$   & $11.07 \pm 7.03$    \\
\textbf{Returns}       & $-16.11 \pm 8.81$ & $-23.06 \pm 12.00$     & $-26.45 \pm 19.25$ & $-13.57 \pm 5.87$  & $-14.41 \pm 6.23$   \\ \midrule
\textbf{}              & \multicolumn{5}{c}{$w=2.0$}                                                                                  \\ \midrule
\textbf{Interventions} & $17.19 \pm 7.42$  & $0.0 \pm 0.0$          & $22.99 \pm 0.12$   & $8.56 \pm 7.47$    & $11.71 \pm 7.16$    \\
\textbf{Returns}       & $-16.11 \pm 8.81$ & $-20.9 \pm 8.03$       & $-26.45 \pm 19.25$ & $-18.40 \pm 13.90$ & $-12.59 \pm 5.13$   \\ \bottomrule
\end{tabular}
\end{table}

\begin{table}[h]
\centering
\caption{\ac{a2c}, pendulum: mean and standard deviation of returns and safeguard interventions for different scaling factors $w$.}
\label{tab:pend_a2c}
\begin{tabular}{@{}lccccc@{}}
\toprule
\textbf{} & \textbf{SE-RL} & \textbf{SE-RL Penalty} & \textbf{SP-RL} & \textbf{SP-RL PSL} & \textbf{SP-RL PenC} \\ \midrule
& \multicolumn{5}{c}{$w=0.1$} \\ \midrule
\textbf{Interventions}      & $170.03 \pm 53.11$   & $1.04 \pm 2.82$ &  $170.03 \pm 53.11$ & $6.06 \pm 8.88$ & $1.04 \pm 2.82$ \\
\textbf{Returns} & $-177.09 \pm 31.28$ & $-32.90 \pm 57.86$ & $-177.09 \pm 31.28$ & $-7.60 \pm 6.33$ & $-32.90 \pm 57.86$ \\ 
\midrule
& \multicolumn{5}{c}{$w=0.5$} \\ \midrule
\textbf{Interventions}      & $170.03 \pm 53.11$  & $0.48 \pm 2.15$ & $170.03 \pm 53.11$  & $7.84 \pm 11.95$ &  $0.48 \pm 2.15$\\
\textbf{Returns} & $-177.09 \pm 31.28$ & $-24.58 \pm 40.57$    & $-177.09 \pm 31.28$ &  $-7.52 \pm 6.54$& $-24.58 \pm 40.57$ \\ 
\midrule
& \multicolumn{5}{c}{$w=1.0$} \\  \midrule
\textbf{Interventions}      & $170.03 \pm 53.11$   & $50.79 \pm 80.19$ & $170.03 \pm 53.11$  & $27.50 \pm 36. 68$ & $50.79 \pm 80.19$ \\
\textbf{Returns} & $-177.09 \pm 31.28$ & $-114.15 \pm 77.99$  & $-177.09 \pm 31.28$ & $-8.02 \pm 7.40$ & $-114.15 \pm 77.99$ \\
\midrule
& \multicolumn{5}{c}{$w=2.0$} \\ \midrule
\textbf{Interventions}     & $170.03 \pm 53.11$   & $38.66 \pm 68.81$ & $170.03 \pm 53.11$  & $11.66 \pm 20.81$ & $38.66 \pm 68.81$ \\
\textbf{Returns} & $-177.09 \pm 31.28$ & $-156.37 \pm 64.00$ & $-177.09 \pm 31.28$ & $-7.06 \pm 6.42$ & $-156.37 \pm 64.00$ \\ 
\bottomrule
\end{tabular}
\end{table}

\begin{table}[h]
\centering
\caption{\ac{a2c}, quadrotor: mean and standard deviation of returns and safeguard interventions for different scaling factors $w$.}
\label{tab:quad_a2c}
\begin{tabularx}{\textwidth}{@{}Xccccc@{}}
\toprule
\textbf{} & \textbf{SE-RL} & \textbf{SE-RL Penalty} & \textbf{SP-RL} & \textbf{SP-RL PSL} & \textbf{SP-RL PenC} \\ \midrule
& \multicolumn{5}{c}{$w=0.1$}\\ \midrule
\textbf{Interventions} & $199.90 \pm 0.46$ & $161.56 \pm 50.16$ & $199.90 \pm 0.46$& $198.27 \pm 4.67$ &$161.56 \pm 50.16$ \\
\textbf{Returns} & $-66.09 \pm 20.09$ & $-50.46 \pm 8.22$ & $-66.09 \pm 20.09$ & $-67.58 \pm 25.27$& $-50.46 \pm 8.22$ \\ 
\midrule
& \multicolumn{5}{c}{$w=0.5$}\\ \midrule
\textbf{Interventions} & $199.90 \pm 0.46$ & $98.27 \pm 64.17$ & $199.90 \pm 0.46$& $198.46 \pm 2.19$ & $98.27 \pm 64.17$\\
\textbf{Returns} & $-66.09 \pm 20.09$ & $-52.75 \pm 12.63$ & $-66.09 \pm 20.09$ & $-58.60 \pm 9.41$& $-52.75 \pm 12.63$\\ 
\midrule
& \multicolumn{5}{c}{$w=1.0$}\\ \midrule
\textbf{Interventions}      &$199.90 \pm 0.46$& $63.96 \pm 63.01$& $199.90 \pm 0.46$ & $196.94 \pm 4.97$ & $63.96 \pm 63.01$\\
\textbf{Returns}   & $-66.09 \pm 20.09$ & $-50.29 \pm 13.41$& $-66.09 \pm 20.09$ & $-62.94 \pm 10.30$ & $-50.29 \pm 13.41$\\ 
\midrule
& \multicolumn{5}{c}{$w=2.0$}\\ \midrule
\textbf{Interventions} & $199.90 \pm 0.46$ &$80.19 \pm 82.43$ & $199.90 \pm 0.46$ & $195.77 \pm 5.66$ & $80.19 \pm 82.43$\\
\textbf{Returns} & $-66.09 \pm 20.09$ &$-51.52 \pm  14.81$ & $-66.09 \pm 20.09$ & $-63.81 \pm 12.06$ & $-51.52 \pm  14.81$\\
\bottomrule
\end{tabularx}
\end{table}

\begin{table}[h]
\centering
\caption{\ac{a2c},  seeker: mean and standard deviation of returns and safeguard interventions for different scaling factors $w$.}
\label{tab:seeker_a2c}
\begin{tabular}{@{}lccccc@{}}
\toprule
\textbf{} & \textbf{SE-RL} & \textbf{SE-RL Penalty} & \textbf{SP-RL} & \textbf{SP-RL PSL} & \textbf{SP-RL PenC} \\ \midrule
\textbf{Interventions} & $56.93 \pm 44.75$   & $66.42 \pm 44.02$  & $56.93 \pm 44.75$   & $43.71 \pm 42.01$      & $66.42 \pm 44.02$      \\
\textbf{Returns}       & $-19.46 \pm 19.92$ & $-22.04 \pm 20.23$ & $-19.46 \pm 19.92$ & $-24.82 \pm 17.54$     & $-22.04 \pm 20.23$      \\ 
\midrule
                       & \multicolumn{5}{c}{$w=0.5$}                                                                                     \\ \midrule
\textbf{Interventions} & $56.93 \pm 44.75$   & $97.43 \pm 15.16$  & $56.93 \pm 44.75$   & $61.33 \pm 40.95$       & $97.43 \pm 15.16$       \\
\textbf{Returns}       & $-19.46 \pm 19.92$ & $-41.33 \pm 15.46$ & $-19.46 \pm 19.92$ & $-35.58 \pm 13.76$     & $-41.33 \pm 15.46$      \\ 
\midrule
                       & \multicolumn{5}{c}{$w=1.0$}                                                                                     \\ \midrule
\textbf{Interventions} & $56.93 \pm 44.75$   & $100.00 \pm 0.00$  & $56.93 \pm 44.75$   & $72.61 \pm 34.33$       & $100.00 \pm 0.00$       \\
\textbf{Returns}       & $-19.46 \pm 19.92$ & $-50.22 \pm 15.04$ & $-19.46 \pm 19.92$ & $-38.61 \pm 13.28$     & $-50.22 \pm 15.04$      \\
\midrule
                       & \multicolumn{5}{c}{$w=2.0$}                                                                                     \\ \midrule
\textbf{Interventions} & $56.93 \pm 44.75$   & $100.00 \pm 0.00$   & $56.93 \pm 44.75$   & $64.86 \pm 36.08$       & $100.00 \pm 0.00$       \\
\textbf{Returns}       & $-19.46 \pm 19.92$ & $-56.06 \pm 14.32$  & $-19.46 \pm 19.92$ & $-40.67 \pm 13.33$     & $-56.06 \pm 14.32$       \\ 
\bottomrule
\end{tabular}
\end{table}

\begin{table}[]
\caption{A2C, EMS: mean and standard deviation of returns and safeguard interventions for different scaling factors $w$.}
\label{tab:ems_a2c}
\begin{tabular}{@{}lccccc@{}}
\toprule
                       & \textbf{SE-RL}   & \textbf{SE-RL Penalty} & \textbf{SP-RL}   & \textbf{SP-RL PSL} & \textbf{SP-RL PenC} \\ \midrule
\textbf{}              & \multicolumn{5}{c}{$w=0.1$}                                                                             \\ \midrule
\textbf{Interventions} & $17.49 \pm 1.67$ & $12.6 \pm 4.02$        & $17.49 \pm 1.67$ & $16.23 \pm 4.46$   & $10.87 \pm 5.51$    \\
\textbf{Returns}       & $-14.14\pm 7.92$ & $-14.12 \pm 7.91$      & $-14.14\pm 7.92$ & $-21.60 \pm 15.51$ & $-13.2 \pm 7.70$    \\ \midrule
\textbf{}              & \multicolumn{5}{c}{$w=0.5$}                                                                             \\ \midrule
\textbf{Interventions} & $17.49 \pm 1.67$ & $7.34 \pm 4.27$        & $17.49 \pm 1.67$ & $13.9 \pm 8.78$    & $0.6 \pm 1.45$      \\
\textbf{Returns}       & $-14.14\pm 7.92$ & $-13.9 \pm 7.84$       & $-14.14\pm 7.92$ & $-15.63 \pm 7.11$  & $-13.01\pm 7.62$    \\ \midrule
\textbf{}              & \multicolumn{5}{c}{$w=1.0$}                                                                             \\ \midrule
\textbf{Interventions} & $17.49 \pm 1.67$ & $4.7 \pm 3.59$         & $17.49 \pm 1.67$ & $15.81 \pm 4.94$   & $0.0 \pm 0.0$       \\
\textbf{Returns}       & $-14.14\pm 7.92$ & $-13.69 \pm 7.79$      & $-14.14\pm 7.92$ & $-13.57 \pm 5.87$  & $-13.23 \pm 7.64$   \\ \midrule
\textbf{}              & \multicolumn{5}{c}{$w=2.0$}                                                                             \\ \midrule
\textbf{Interventions} & $17.49 \pm 1.67$ & $2.16 \pm 2.59$        & $17.49 \pm 1.67$ & $8.56 \pm 7.47$    & $0.0 \pm 0.0$       \\
\textbf{Returns}       & $-14.14\pm 7.92$ & $-13.38 \pm 7.82$      & $-14.14\pm 7.92$ & $-18.40 \pm 13.90$ & $-13.4 \pm 7.65$    \\ \bottomrule
\end{tabular}
\end{table}
\end{document}